\documentclass[preprint,12pt,authoryear]{elsarticle}

\usepackage{amssymb}
\usepackage{amsmath}
\usepackage{url}           
\usepackage{hyperref}
\usepackage{dsfont}
\usepackage{booktabs}
\usepackage{multirow}
\usepackage{float}
\usepackage{algorithm}
\usepackage{xcolor}
\usepackage{algorithmic}
\usepackage[normalem]{ulem}
\definecolor{mygray}{gray}{0.5}

\hypersetup{
    colorlinks=true,
    citecolor=blue,
    linkcolor=red,
    filecolor=magenta,      
    urlcolor=cyan,
}

\journal{Neural Networks}

\begin{document}

\begin{frontmatter}
\title{BCMDA: Bidirectional Correlation Maps Domain Adaptation for Mixed Domain Semi-Supervised Medical Image Segmentation}

\author[label1]{Bentao Song} 
\ead{jzjaha@mails.swust.edu.cn}
\author[label1]{Jun Huang}
\ead{huangjuncs@swust.edu.cn}
\author[label1]{Qingfeng Wang\corref{cor1}}
\ead{qfwang@swust.edu.cn}
\cortext[cor1]{Corresponding author}

\address[label1]{School of Computer Science and Technology, Southwest University of Science and Technology, Mianyang, Sichuan, 621010, China}

\begin{abstract}
In mixed domain semi-supervised medical image segmentation (MiDSS), achieving superior performance under domain shift and limited annotations is challenging. This scenario presents two primary issues: (1) distributional differences between labeled and unlabeled data hinder effective knowledge transfer, and (2) inefficient learning from unlabeled data causes severe confirmation bias. In this paper, we propose the bidirectional correlation maps domain adaptation (BCMDA) framework to overcome these issues. On the one hand, we employ knowledge transfer via virtual domain bridging (KTVDB) to facilitate cross-domain learning. First, to construct a distribution-aligned virtual domain, we leverage bidirectional correlation maps between labeled and unlabeled data to synthesize both labeled and unlabeled images, which are then mixed with the original images to generate virtual images using two strategies, a fixed ratio and a progressive dynamic MixUp. Next, dual bidirectional CutMix is used to enable initial knowledge transfer within the fixed virtual domain and gradual knowledge transfer from the dynamically transitioning labeled domain to the real unlabeled domains. On the other hand, to alleviate confirmation bias, we adopt prototypical alignment and pseudo label correction (PAPLC), which utilizes learnable prototype cosine similarity classifiers for bidirectional prototype alignment between the virtual and real domains, yielding smoother and more compact feature representations. Finally, we use prototypical pseudo label correction to generate more reliable pseudo labels. Empirical evaluations on three public multi-domain datasets demonstrate the superiority of our method, particularly showing excellent performance even with very limited labeled samples. Code available at \url{https://github.com/pascalcpp/BCMDA}.
\end{abstract}

\end{frontmatter}

\section{Introduction}
\label{introduction}
Although semi-supervised medical image segmentation (SSMS) has demonstrated its potential in scenarios with limited annotated data~\citep{yu2019uncertainty,chi2024adaptive,ZHANG2024110426,miao2023caussl}, and some SSMS methods have even surpassed the upper bound performance achieved with fully annotated data~\citep{bai2023bidirectional,ZHANG2025111722,song2024sdcl,WANG2025111116}, a critical limitation persists: SSMS methods typically assume that labeled and unlabeled data follow the same distribution, originating from the same domain, medical center, or imaging scanner~\citep{bai2017semi,yu2019uncertainty}. However, in real-world scenarios, this assumption is often violated, as labeled data are typically obtained from a limited number of institutions due to the high cost and expertise required for manual annotation, whereas unlabeled data are more easily collected from multiple medical centers using heterogeneous imaging devices and acquisition protocols~\citep{guan2021domain}. These inter-institutional and cross-scanner discrepancies introduce substantial domain gaps, which severely hinder effective knowledge transfer from labeled to unlabeled data and make stable and accurate pseudo-label generation particularly challenging. This SSMS task with domain shift~\citep{guan2021domain} is referred to as mixed domain semi-supervised medical image segmentation (MiDSS)~\citep{ma2024constructing}. MiDSS presents two primary challenges: (1) domain shift caused by distributional differences between labeled and unlabeled data, and (2) the scarcity of labeled data. Therefore, given the limitations of traditional SSMS methods, there is an urgent need for alternative, concise, and effective approach in MiDSS.

\begin{figure}[!t]
\centerline{\includegraphics[width=1\textwidth]{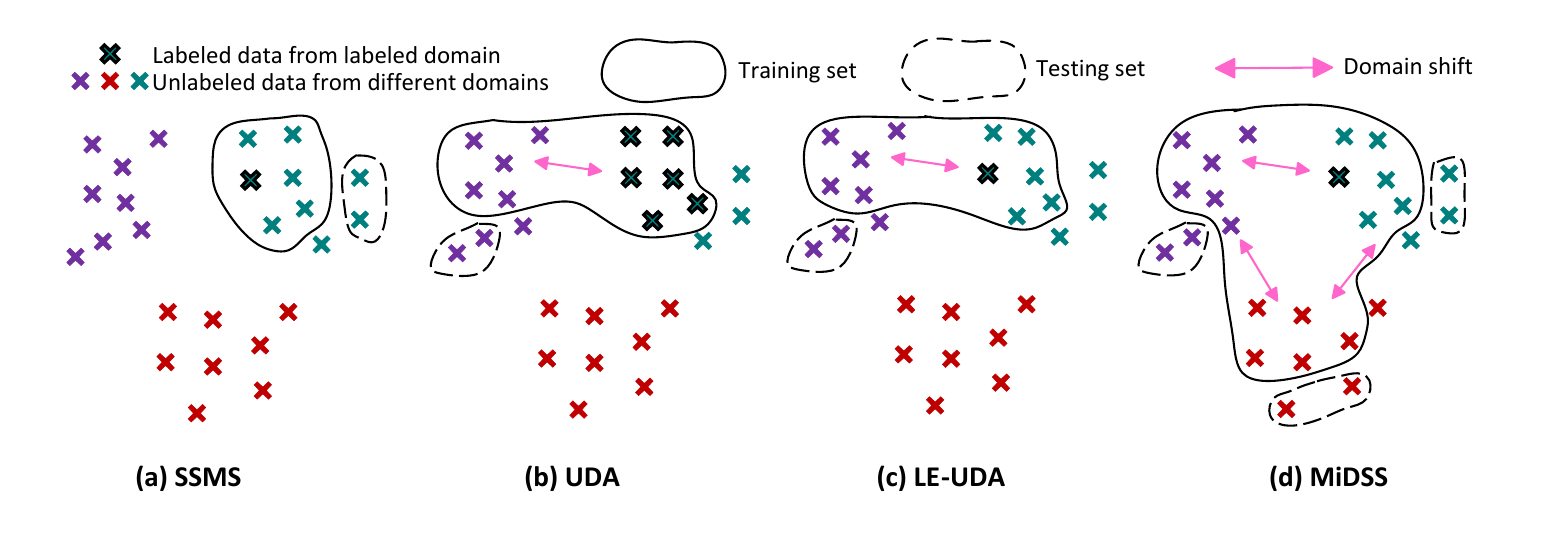}}
\vspace{-4mm}
\caption{From (a) to (d), the divisions of the training and testing sets represent different scenario settings, with the difficulty increasing sequentially.}
\label{fig1}
\end{figure}

To address these challenges, particularly domain shift, unsupervised domain adaptation (UDA)~\citep{chen2020unsupervised,chartsias2017adversarial,huy2022adversarial} methods have shown promising potential. However, UDA typically assumes an abundance of labeled data in the source domain, making it unsuitable for the scenario considered in this study. Additionally, a recent study closely related to our scenario, known as label-efficient unsupervised domain adaptation (LE-UDA)~\citep{zhao2022uda}, considers a setting where labeled data is limited and originates from a single domain, while unlabeled data includes samples from both the source domain and the target domain. Although LE-UDA appears to address the challenges discussed in this study, it is important to note that it is designed specifically for scenarios with a single unlabeled target domain and does not provide a solution for cases involving multiple unlabeled target domains. Fig.~\ref{fig1} illustrates the differences between the aforementioned scenarios. Effectively leveraging the unlabeled data from multiple target domains, along with the limited labeled data, remains an important research challenge. 

\begin{figure}[!t]
\centerline{\includegraphics[width=1.0\textwidth]{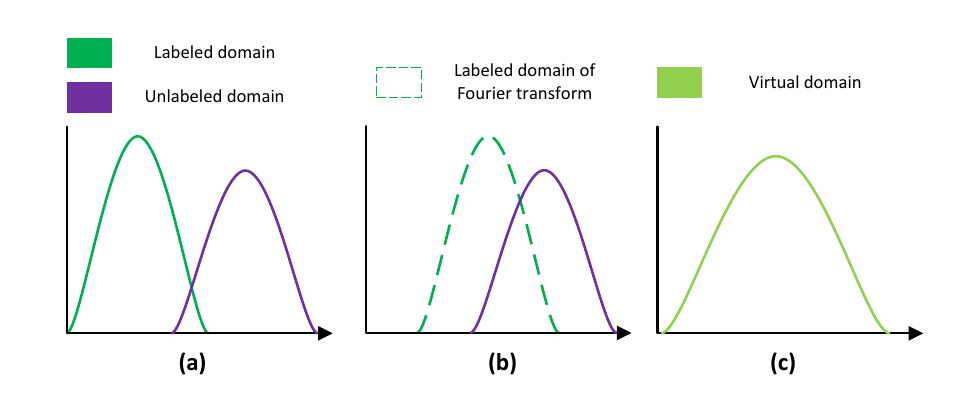}}
\vspace{-4mm}
\caption{Illustration of distribution differences under MiDSS: (a) the original distribution of labeled and unlabeled data, (b) labeled data distribution after a unidirectional transformation used in SymGD, and (c) the well-aligned virtual domain distribution formed by both labeled and unlabeled data after the bidirectional transformation with our method.}
\label{intro_fig2}
\end{figure}

SymGD~\citep{ma2024constructing} serves as an initial attempt to address the MiDSS task, but it still has significant limitations. \textbf{First}, it employs a unidirectional Fourier transformation on the labeled data and applies a unified copy-paste (UCP) with the unlabeled data to construct several intermediate domains for knowledge transfer. Nevertheless, the transformed labeled data and the unlabeled data may still exhibit substantial differences, hindering effective knowledge transfer within the intermediate domains. \textbf{Second}, considering the inherent confirmation bias~\citep{song2024sdcl} in SSMS,
this issue becomes even more severe in MiDSS, leading to increasingly unstable and erroneous pseudo labels—a problem that SymGD has not explored. Consequently, the accumulation of confirmation bias during training results in severe cognitive bias~\citep{song2024sdcl} in the model, which is one of the primary factors hindering its performance. In summary, the limitations of the above methods lead to suboptimal results, highlighting the need for further refinement in domain bridging techniques.

In this work, we propose a framework named bidirectional correlation maps domain adaptation (BCMDA) to address the issues of domain shift and confirmation bias. As illustrated in Fig.~\ref{intro_fig2}, BCMDA effectively aligns the distributions of labeled and unlabeled data, significantly mitigating the domain shift problem. Specifically, we introduce a method called knowledge transfer via virtual domain bridging (KTVDB). Inspired by CorrMatch~\citep{sun2024corrmatch}, which generates new predictions through propagation using correlation maps in the output space, we explore the feasibility of applying this approach in the input space as well. KTVDB leverages bidirectional correlation maps (BCM) between labeled and unlabeled data to guide the synthesis of images from one domain using pixels from another domain, and then mixes them with the original images through the fixed-ratio MixUp (FixMix) strategy. This process incorporates the pixel distribution of a different domain while preserving the original distribution, resulting in the formation of a virtual domain distribution. By applying this operation to both labeled and unlabeled data, their distributions are aligned with the virtual domain distribution, respectively. Through bidirectional distribution alignment, virtual labeled and virtual unlabeled data collectively form a distribution-aligned virtual domain.

To facilitate a gradual stylistic transition between the real domains, we apply the progressive dynamic MixUp (PDMix) to mix the synthesized labeled data and real labeled data. This generates dynamic virtual labeled data that smoothly transfers the distribution from the labeled domain to the unlabeled domains. Subsequently, we employ a bidirectional CutMix (BCMix) technique twice, which similar to bidirectional copy-paste (BCP)~\citep{bai2023bidirectional}. Dual BCMix enables efficient initial knowledge transfer between virtual labeled data and virtual unlabeled data, and further facilitates smooth knowledge transfer from the labeled domain to the unlabeled domains via interactions between the dynamic virtual labeled data and the real unlabeled data. 

Moreover, some methods~\citep{feng2023unsupervised,zhang2021prototypical} exploit class prototypes to aggregate semantically similar features across domains and denoise pseudo labels, while others~\citep{lu2020stochastic,zhou2023semi} employ stochastic classifiers to identify additional misaligned local regions, thereby alleviating overfitting. Inspired by the observation that the weights of stochastic classifiers can serve as class prototypes~\citep{zhou2023semi}, we eliminate their stochasticity and develop a stable and learnable prototype cosine similarity (CosSim) classifier. Based on this, we propose a prototypical alignment and pseudo label correction (PAPLC) approach, which uses the CosSim classifier to simultaneously facilitate prototype extraction and feature alignment during training. Furthermore, considering the initial feature discrepancies between the virtual and real domains, we perform bidirectional prototype alignment (BPA) to progressively obtain well-aligned feature representations. As the model inevitably produces linearly inseparable feature vectors~\citep{zhang2021prototypical}, we design prototypical pseudo label correction (PPLC), which utilizes the predictions of the CosSim classifier to correct the predictions of the Linear classifier, effectively mitigating confirmation bias.

In conclusion, the contributions of this paper can be summarized as follows:
\begin{itemize}
    \item We propose a novel domain adaptation framework that addresses the domain shift problem by utilizing BCM for bidirectional data distribution alignment. 
    \item We present KTVDB, which utilizes data synthesized through BCM to perform FixMix and PDMix, respectively, and further applies dual BCMix (DBCMix) to enable efficient knowledge transfer via virtual domain bridging.
    \item We propose PAPLC, which adopts BPA based on the CosSim classifiers to simultaneously achieve comprehensive and stable prototype acquisition and feature alignment. Additionally, PPLC leverages CosSim classifier to correct the predictions of Linear classifier, significantly alleviating confirmation bias.   
    \item Extensive experiments on three public multi-domain datasets demonstrate the superiority of the proposed method. Our approach outperforms other state-of-the-art (SOTA) methods across all three datasets.
\end{itemize} 

\section{Related Work}
\subsection{Semi-supervised medical image segmentation}
\label{II-A}
Semi-supervised medical image segmentation (SSMS) has demonstrated significant potential in addressing the challenge of scarce annotations in medical image segmentation tasks and has therefore attracted extensive research attention. Most SSMS methods can be broadly categorized into several groups, including pseudo-labeling, consistency regularization, data augmentation–based approaches, and other methodological paradigms.

The purpose of pseudo-labeling~\citep{bai2017semi,luo2022semi,song2024sdcl} is to generate pseudo labels for unlabeled data, which are then used as supervisory signals to train the model. FixMatch~\citep{sohn2020fixmatch} generates pseudo labels from weakly augmented images and employs them to supervise the corresponding strongly augmented samples after discarding unreliable predictions, demonstrating strong performance despite its conceptual simplicity. PEFAT~\citep{zeng2023pefat} splits a high-quality pseudo-labeled set based on the pseudo-loss and performs feature-level adversarial training.

Consistency-based~\citep{luo2021semi,luo2022semiurpc,miao2023caussl} SSMS methods generally train models by enforcing prediction consistency on unlabeled data under different perturbations. Mean Teacher (MT) framework~\citep{tarvainen2017mean} constructs a teacher model using the exponential moving average (EMA) of the student model and enforces consistency between their predictions. UA-MT~\citep{yu2019uncertainty} filters out unreliable predictions by estimating uncertainty. UPCoL~\citep{lu2023upcol} leverages uncertainty-guided prototype consistency.~\cite{lv2025clustering} proposes clustering-guided contrastive prototype learning to alleviate class imbalance. LeFeD~\citep{zeng2024consistency} improves performance by enlarging and exploiting feature-level discrepancies between dual decoders.

Data augmentation–based methods~\citep{bai2023bidirectional,chi2024adaptive} are essentially an extension of consistency regularization. DACNet~\citep{WANG2025111116} designs dedicated data augmentation strategies together with a dual attention–guided consistency network, while~\cite{wang2025dynamic} proposes a dynamic mask stitching–guided region consistency network, both of which effectively mitigate cognitive bias. PICK~\citep{zeng2025pick} exploits unlabeled data by masking and predicting pseudo-label–guided attentive regions.

In addition, there are several SSMS methods that explore alternative perspectives. VerSemi~\citep{zeng2025segment} leverages more unlabeled data by integrating multiple tasks into a unified model. TextMoE~\citep{zeng2025exploring} explores the potential of jointly learning from related text–image datasets, while TeViA~\citep{zeng2025harnessing} proposes a segmentation-specific text-to-vision alignment framework.

Although the aforementioned methods have extensively studied SSMS scenario, these SSMS methods typically assume no domain shift~\citep{bai2017semi,yu2019uncertainty} between labeled and unlabeled data. Therefore, conducting research in scenarios with domain shift is essential.

\subsection{Unsupervised medical domain adaptation}
\label{II-B}

In real-world scenarios, medical image segmentation models often face domain shifts due to factors like imaging device differences or patient-specific anatomical variations. Given the high annotation cost of medical images, unsupervised domain adaptation (UDA) has gained attention by enabling model adaptation to target domains with only unlabeled data. Recent UDA methods aim to reduce domain discrepancies by constructing various intermediate domains~\citep{NA2025111537,xu2020adversarial,yang2020fda,chen2022deliberated}, progressively narrowing the domain gap. This mechanism, known as domain bridging (DB), aligns the distributions of the source and target domains to form intermediate representations. Among these, style transfer-based DB methods are the most common. These methods typically leverage generative models, such as Generative Adversarial Networks (GANs)~\citep{chartsias2017adversarial,zhao2022uda} or diffusion models~\citep{gong2024diffuse,benigmim2023one}, to transform the style of source domain data into that of the target domain, or vice versa, aiming to mitigate the domain gap. However, these methods often suffer from high computational complexity, difficulty in converging, and may introduce unexpected artifacts in the generated images, leading to semantic ambiguity.

Another category of DB methods involves global interpolation~\citep{zhang2018mixup, NA2025111537} and local replacement~\citep{yun2019cutmix,chen2022deliberated}. Global interpolation methods, such as MixUp~\citep{zhang2018mixup}, generally perform data mixing but may cause pixel-level semantic ambiguity. In contrast, local replacement methods, such as CutMix~\citep{yun2019cutmix}, tend to preserve semantic consistency more effectively. However, combining patches from different distributions can still introduce stylistic differences~\citep{ma2024constructing}, which hinder effective knowledge transfer. Additionally, methods based on Fourier transforms~\citep{yang2020fda, ma2024constructing} align distributions by swapping the low-frequency spectrum between images. Yet, hard replacement of the low-frequency spectrum from other data may introduce additional noise. Furthermore, the transformation primarily focuses on visual style, exerting limited influence on the actual pixel distribution of the image.

UDA methods can also be broadly categorized into input-level, feature-level, and output-level alignment approaches. Input-level alignment methods primarily rely on image translation or style manipulation techniques, including CycleGAN-based image translation~\citep{bousmalis2017unsupervised,chartsias2017adversarial}, Fourier domain adaptation~\citep{yang2020fda}, diffusion-based data synthesis~\citep{gong2024diffuse,benigmim2023one}, as well as global interpolation~\citep{xu2020adversarial,NA2025111537} and local replacement strategies~\citep{chen2022deliberated,zhou2022context} to bridge the domain gap.. Feature-level alignment methods aim to learn domain-invariant representations by aligning feature distributions across domains, typically through adversarial learning~\citep{ganin2016domain}, content-style disentanglement~\citep{huy2022adversarial}, or unified frameworks that jointly consider input- and feature-level alignment~\citep{chen2020unsupervised}. Output-level alignment methods focus on minimizing the discrepancy between prediction distributions of different domains via adversarial learning~\citep{tsai2018learning,chen2022reusing}.

In recent work,~\cite{zhao2022uda} proposed LE-UDA, combining input and feature alignment with inter- and intra-domain knowledge transfer to achieve strong performance.~\cite{ma2024constructing} later extended LE-UDA to multiple target domains, introducing the MiDSS scenario with a SymGD bridging strategy. However, the Fourier transformed labeled data in SymGD may still mismatch unlabeled data distributions, and the confirmation bias issue in pseudo-labeling is overlooked. These limitations significantly impact model performance, motivating our work. This study achieves strong performance by unifying the MixUp, CutMix, and FixMatch paradigms from SSMS and UDA, while proposing the KTVDB and PAPLC components.

\section{Methodology}
\subsection{Problem definition}
\label{III-A}
Unlike traditional SSMS and UDA tasks, the MiDSS task involves both labeled data from a single domain and unlabeled data from all domains. Specifically, there exists \( \mathcal{D} = \{\mathcal{D}_1, \mathcal{D}_2, \dots, \mathcal{D}_K\} \), where \( K \) represents the number of domains. The labeled dataset \( \mathcal{D}^l = \{(x_i, y_i)\}_{i=1}^N \) comes from a single domain \( \mathcal{D}_j \in \mathcal{D} \) with \( N \) labeled images, while the unlabeled dataset \( \mathcal{D}^u = \{u_i\}_{i=1}^M \) consists of \( M \) unlabeled images from all domains, where \( M \gg N \). Each 2D medical image \( x_i, u_i \in \mathbb{R}^{D \times H \times W} \), where \( D \) and \( H \) and \( W \) represent the channel and height and width, respectively. For \( \mathcal{D}^l \), the label \( y_i \in \{0,1\}^{C \times H \times W} \), where \( C \) denotes the number of classes. We aim to leverage the limited labeled data from a single labeled domain and the abundant unlabeled data from all domains to train a model that performs well across all domains.



\subsection{Overall architecture}
\label{III-B}
\begin{figure}[!t]
\centerline{\includegraphics[width=\columnwidth]{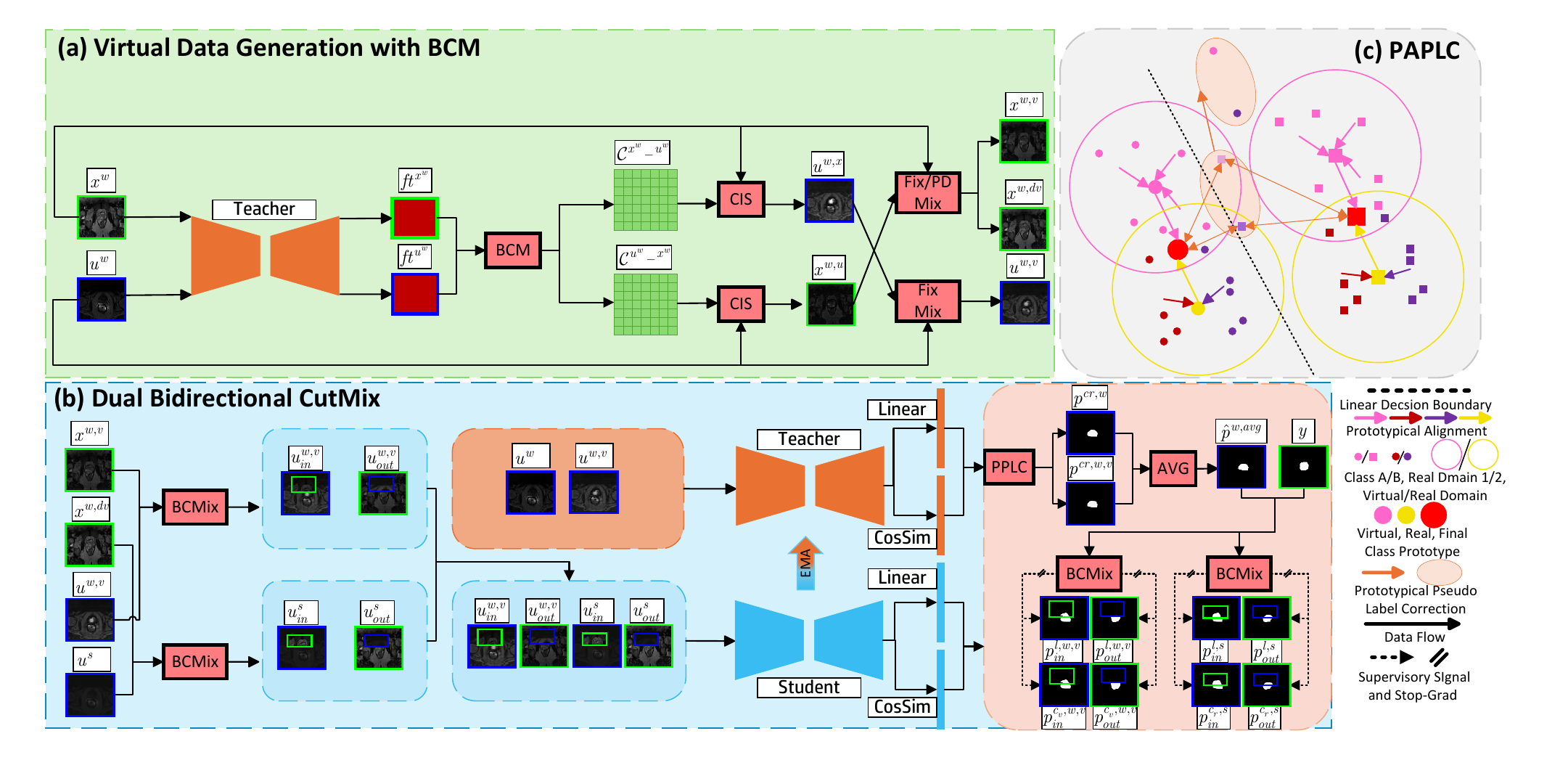}}
\vspace{-4mm}

\caption{This figure illustrates the overall architecture of bidirectional correlation maps domain adaptation (BCMDA). (a) The specific process of virtual data generation, where CIS stands for the correlation map-informed image synthesis operation. (b) Mixed training data is generated using dual bidirectional CutMix, enabling efficient knowledge transfer by leveraging the virtual domain as a bridge, where AVG refers to the operation of averaging two probability maps. (c) The prototypical alignment and pseudo label correction (PAPLC) component, which employs bidirectional prototype alignment for smooth feature aggregation. Additionally, for classification results erroneously assigned to the class on the right due to a linear decision boundary, it is observed that their features are closer to the prototype on the left. In such cases, the predictions are corrected to the left-side class.}
\label{framework}
\end{figure}

Fig.~\ref{framework} illustrates the overall architecture of the BCMDA framework, which consists of the KTVDB component and the PAPLC component. The KTVDB component first generates both fixed and dynamic virtual data, then performs effective knowledge transfer through dual bidirectional CutMix, while the PAPLC component aligns features and mitigates confirmation bias.

Our framework is based on the mean teacher (MT) architecture~\citep{tarvainen2017mean}, where \( f^s \) and \( f^t \) denote the backbones of the student and teacher networks, respectively, while \( g^s \) and \( g^t \) represent their corresponding classifiers. The parameters of \( f^t \) and \( g^t \) are updated via an exponential moving average (EMA) of \( f^s \) and \( g^s \). Given an image \( x \), its weakly augmented version is denoted as \( x^w = \mathcal{A}^w(x) \), while \( u^w \) represents the weakly augmented unlabeled image. Furthermore, \( u^s = \mathcal{A}^s(u^w) \) denotes the strongly augmented variant of \( u^w \). The feature representation of the weakly augmented unlabeled image, denoted as \( f^t(u^w) \), is obtained through the teacher backbone, while the corresponding probability map and pseudo label are computed as follows:  
\begin{equation}
   ft^{u^w} = f^t(u^w); p^w = g^t(f^t(u^w)); \hat{p} = \arg\max(p^w), 
\end{equation}
where \( ft^{u^w} \) represents the extracted feature of the unlabeled image, \( p^w \) denotes the probability map, and \( \hat{p} \) is obtained by applying the \( \text{argmax} \) operation on \( p^w \), resulting in a pseudo label with a one-hot probability distribution. Similarly, the feature representation of the labeled image, denoted as \( ft^{x^w} \), can be derived in the same manner.

\subsection{Knowledge transfer via virtual domain bridging}
\label{III-C}
\subsubsection{Virtual data generation with BCM}
To generate the bidirectional correlation maps (BCM), we first obtain the feature representations \( ft^{u^w} \) and \( ft^{x^w} \in \mathbb{R}^{D' \times H \times W} \) from the teacher backbone, where \( D' \) denotes the number of feature channels. To reduce computational complexity, we apply downsampling to the feature maps, followed by a reshaping operation, yielding \( ft^{{u^w}, down} \) and \( ft^{{x^w}, down} \in \mathbb{R}^{D' \times W'W'} \), where \( W' \) represents the spatial resolution after downsampling, and \( W'W' \) denotes the number of feature vectors. Subsequently, we compute the pairwise similarity between pixels to obtain the BCM using the following formula:
\begin{equation}
\label{eqcorr}
\begin{aligned}
    \mathcal{C}^{{x^w}\_{u^w}} &= \operatorname{Softmax}({(ft^{{x^w}, down})}^T \cdot {ft^{{u^w}, down}}/ \sqrt{D'}), \\
    \mathcal{C}^{{u^w}\_{x^w}} &= \operatorname{Softmax}({(ft^{{u^w}, down})}^T \cdot {ft^{{x^w}, down}}/ \sqrt{D'}),
\end{aligned}
\end{equation}
where \( T \) denotes the matrix transpose operation, \( \mathcal{C}^{x^w\_u^w} \) represents the correlation map from labeled to unlabeled data, whereas \( \mathcal{C}^{u^w\_x^w} \) corresponds to the reverse correlation map. Specifically, \( \mathcal{C} \in \mathbb{R}^{W'W' \times W'W'} \) is the result of the matrix multiplication between feature maps followed by softmax normalization, representing the pairwise similarities between pixels. More concretely, in \( \mathcal{C}^{x^w\_u^w} \), each column denotes the correlations between all pixels in the labeled image and a single pixel in the unlabeled image. \begin{figure}[!t]
\centerline{\includegraphics[width=0.60\columnwidth]{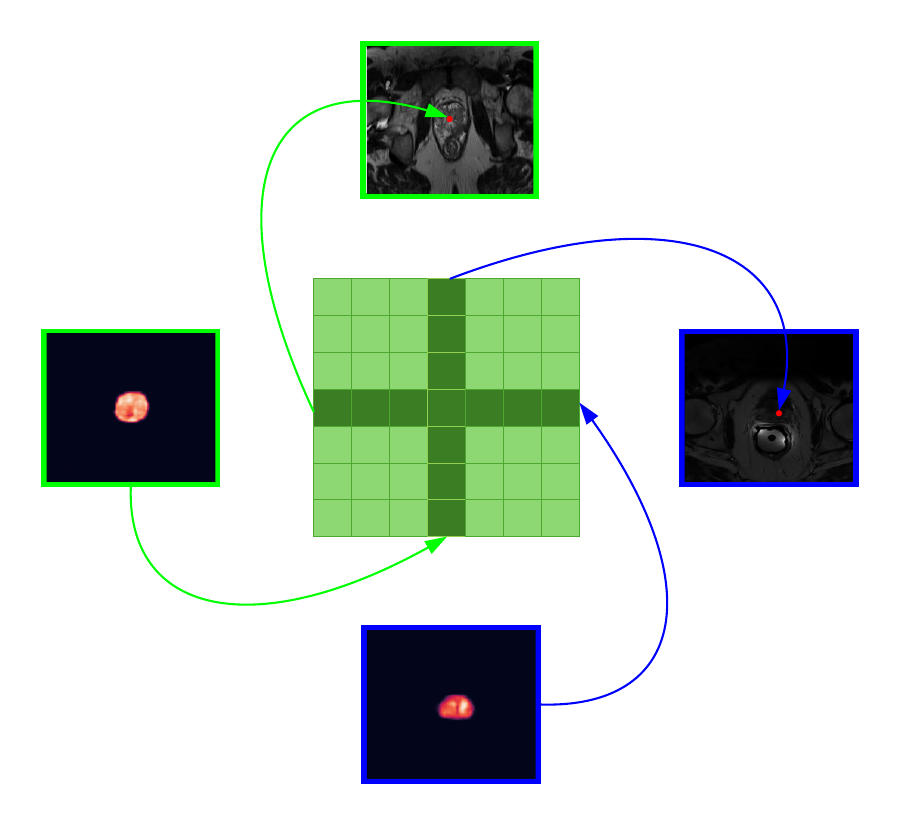}}
\vspace{-4mm}
\caption{Illustration of the bidirectional correlation maps, where the green border and blue border denote the labeled and unlabeled images, respectively, and the correlation map in the middle represents \( \mathcal{C}^{x^w\_u^w} \). Here, we ignore the softmax operation, so \( \mathcal{C}^{u^w\_x^w} \) is simply the transpose of \( \mathcal{C}^{x^w\_u^w} \). From the figure, we can see that each column shows the correlations between all labeled pixels and a single unlabeled pixel, while each row shows the correlations between all unlabeled pixels and a single labeled pixel.}
\label{corrvis}
\end{figure} Fig.~\ref{corrvis} provides a detailed explanation of bidirectional correlation maps. Next, by using the corresponding correlation map, we can synthesize the labeled image using unlabeled pixels and vice versa. By this correlation map-informed image synthesis (CIS) method, we obtain the synthesized labeled and unlabeled images, thereby enabling the exchange of data distributions between labeled and unlabeled data. The corresponding formulas are as follows:
\begin{equation}
\label{eqsythe}
\begin{aligned}
    u^{w,x} &= x^{w, down} \cdot \mathcal{C}^{x^w\_u^w}, \\
    x^{w,u} &= u^{w, down} \cdot \mathcal{C}^{u^w\_x^w},
\end{aligned}
\end{equation}
where \( x^{w, down} \) and \( u^{w, down} \in \mathbb{R}^{D \times W'W'} \) represent the downsampled and reshaped images, while \( x^{w,u} \) and \( u^{w,x} \in \mathbb{R}^{D \times W'W'} \) denote the synthesized labeled and unlabeled images, respectively. For the subsequent steps, we need to reshape and upsample \( x^{w,u} \) and \( u^{w,x} \) to match the original data shape, i.e., \( x^{w,u}, u^{w,x} \in \mathbb{R}^{D \times H \times W} \). 
Although the distribution of the synthesized images has already shifted towards that of the source pixels, achieving a distributional exchange effect between labeled and unlabeled data, their distributions are still not aligned. Moreover, considering the model's incomplete learning introduces feature errors, this results in an inaccurate correlation map, leading to noise artifacts and semantic inconsistencies in the synthesized images. To align the distribution and refine the texture details, we apply the fixed ratio MixUp (FixMix) between the synthesized images and their corresponding real images, generating the final virtual domain images. These virtual labeled and unlabeled images simultaneously retain both their real pixel distribution and source pixel distribution, constructing a distribution-aligned virtual domain. Through this approach, we obtain the data in the virtual domain: 
\begin{equation}
\label{eqa4}
\begin{aligned}
    x^{w,v} &= (1-\lambda_{\text{fix}}) x^{w} + \lambda_{\text{fix}} x^{w, u}, \\
    u^{w,v} &= (1-\lambda_{\text{fix}}) u^{w} + \lambda_{\text{fix}} u^{w, x},
\end{aligned}
\end{equation}
where \( \lambda_{\text{fix}} \) represents the mixing ratio of FixMix, and \( x^{w,v} \) and \( u^{w,v} \) denote the virtual labeled and unlabeled images in the virtual domain, respectively.

Furthermore, to more smoothly utilize the virtual domain as a bridge for progressively transferring labeled data knowledge to unlabeled data, we also construct dynamic virtual labeled data using the progressive dynamic MixUp (PDMix) strategy, which is formulated as: 
\begin{equation}
\label{eqa5}
\begin{aligned}
    x^{w,dv} &= (1-\lambda_{\text{dyn}}) x^{w} + \lambda_{\text{dyn}} x^{w, u}, 
\end{aligned}
\end{equation}
where \( x^{w,dv} \) represents the dynamic virtual labeled data, and \( \lambda_{\text{dyn}} \) denotes the dynamic MixUp ratio. In each iteration, we first sample a value \( \lambda_{\text{dyn}}' \) from a Beta distribution, i.e., \( \lambda_{\text{dyn}}' \sim \text{Beta}(\alpha, \alpha) \), where \( \alpha \) is the parameter of the Beta distribution. To adapt \( \lambda_{\text{dyn}} \) based on the training progress, we apply a scaling factor, leading to \( \lambda_{\text{dyn}} = \gamma \lambda_{\text{dyn}}'\). Here, \( \gamma = \min(\lambda_{\text{fix}}, t/t_{\max}) \) serves as the upper bound of MixUp, dynamically adjusting throughout the training process, where \( t \) and \( t_{\max} \) denote the current and maximum iteration counts, respectively.

Through the aforementioned process, we obtain distribution-aligned virtual labeled and unlabeled data, as well as dynamically evolving virtual labeled data that progressively adapts to the distribution of the real unlabeled data. Subsequently, we employ two simple bidirectional CutMix (BCMix) operations to efficiently transfer knowledge from the labeled data to the unlabeled data.

\begin{figure}[!t]
\centerline{\includegraphics[width=\columnwidth]{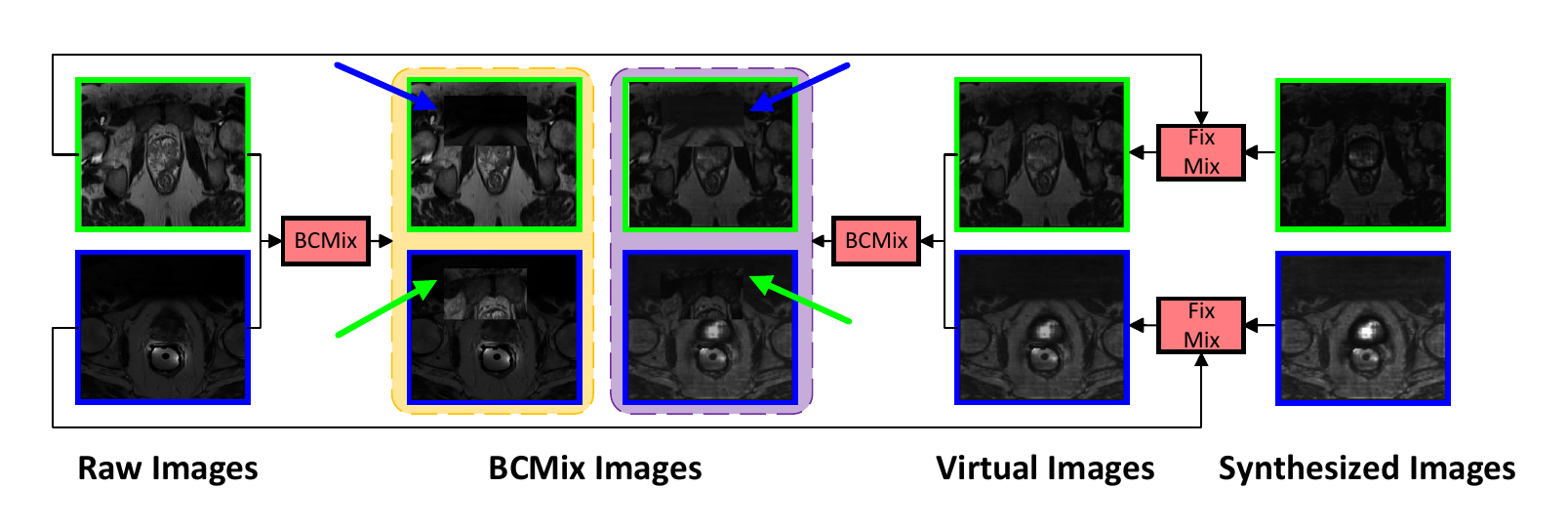}}
\vspace{-4mm}
\caption{This figure compares BCMix applied to raw images with the domain gap and to the distribution-aligned virtual images. On the left, BCMix is applied to real domain images, where the pasted patches exhibit stylistic differences from the backgrounds, hindering knowledge transfer. On the right, BCMix operates on virtual domain images, enabling smooth transitions between the patches and backgrounds for near seamless integration and enhanced knowledge transfer.}
\label{mixvis}
\end{figure}

\subsubsection{Dual bidirectional CutMix}
Applying bidirectional copy-paste (BCP)~\citep{bai2023bidirectional} between labeled and unlabeled data provides a simple yet effective knowledge transfer strategy. Based on this method, we utilize a CutMix mask generation technique that is better suited for handling differences in distribution and semantic context. Specifically, by performing bidirectional CutMix (BCMix) between $x^{w,v}$ and $u^{w,v}$, which are derived from Eq.~\eqref{eqa4}, as well as between $x^{w,dv}$, obtained from Eq.~\eqref{eqa5}, and $u^{s}$, we generate the dual BCMix-augmented images:
\begin{equation}
\label{eqa6}
\begin{alignedat}{4}
    &u^{w,v}_{in} &&= x^{w,v} \odot M_1 + u^{w,v} \odot (\mathbf{1}-M_1), \quad 
    &&u^{w,v}_{out} &&= u^{w,v} \odot M_1 + x^{w,v} \odot (\mathbf{1}-M_1), \\
    &u^{s}_{in} &&= x^{w,dv} \odot M_2 + u^{s} \odot (\mathbf{1}-M_2), \quad 
    &&u^{s}_{out} &&= u^{s} \odot M_2 + x^{w,dv} \odot (\mathbf{1}-M_2), \\
\end{alignedat}
\end{equation}
where $M_1, M_2 \in \{0,1\}^{W \times H}$ are randomly generated masks, where the region with a value of 1 indicates the area for copy-paste. $\mathbf{1}$ represents an all-one matrix with the corresponding shape, and $\odot$ denotes element-wise multiplication.
Through this approach, the dual bidirectional CutMix (DBCMix) effectively and efficiently transfers real labeled knowledge to real unlabeled data. Fig.~\ref{mixvis} highlights the advantages of virtual domain bridging compared to direct knowledge transfer in real domains. 
Algorithm~\ref{alg:KTVDB} shows the pseudocode of KTVDB.

\begin{algorithm*}[!t]
\caption{Pseudocode of KTVDB}
\label{alg:KTVDB}
\begin{algorithmic}[1]
\footnotesize
\STATE {\bfseries Input:}
Labeled dataset $\mathcal{D}^l$;
unlabeled dataset $\mathcal{D}^u$.

    \STATE \hspace{0.90em} Fetch labeled batch $(x,y) \gets \mathcal{D}^l$ and unlabeled batch $u\gets \mathcal{D}^u$

    \STATE\hspace{0.90em} $x^w = \mathcal{A}^w(x)$, $u^w = \mathcal{A}^w(u)$, $u^s = \mathcal{A}^s(u^w)$ \COMMENT{\textit{Weak/Strong augmentation}} 
    
    \STATE\hspace{0.90em} $ft^{u^w} = f^t(u^w)$, $ft^{x^w} = f^t(x^w)$ \COMMENT{\textit{Obtain feature maps}}
    
    \STATE\hspace{0.90em} Generate bidirectional correlation maps $\mathcal{C}^{{x^w}\_{u^w}}$ and $\mathcal{C}^{{u^w}\_{x^w}}$ by Eq.~\eqref{eqcorr}
    
    \STATE\hspace{0.90em} Synthesize images $x^{w,u}$ and $u^{w,x}$ by Eq.~\eqref{eqsythe}

    \STATE\hspace{0.90em} Apply FixMix using Eq.~\eqref{eqa4} to obtain virtual images $x^{w,v}$ and $u^{w,v}$
        \\
    \hspace{0.90em}\COMMENT{\textit{Obtain distribution-aligned virtual domain images}}
    
    \STATE\hspace{0.90em} Apply PDMix using Eq.~\eqref{eqa5} to obtain dynamic virtual images $x^{w,dv}$
        \\
    \hspace{0.90em}\COMMENT{\textit{Obtain dynamic virtual labeled images}}

    \STATE\hspace{0.90em} Randomly generate masks $M_1$ and $M_2$

    \STATE\hspace{0.90em} $u^{w,v}_{in} = x^{w,v} \odot M_1 + u^{w,v} \odot (\mathbf{1}-M_1), u^{w,v}_{out} = u^{w,v} \odot M_1 + x^{w,v} \odot (\mathbf{1}-M_1)$ 
    \\
    \hspace{0.90em}\COMMENT{\textit{Initial knowledge transfer in the fixed virtual domain}}

    \STATE\hspace{0.90em} $u^{s}_{in} = x^{w,dv} \odot M_2 + u^{s} \odot (\mathbf{1}-M_2), u^{s}_{out} = u^{s} \odot M_2 + x^{w,dv} \odot (\mathbf{1}-M_2)$ 
    \\
    \hspace{0.90em}\COMMENT{\textit{Knowledge transfer from the dynamic labeled domain to unlabeled domains}}

\RETURN $u^{w,v}_{in},u^{w,v}_{out},u^{s}_{in},u^{s}_{out}$
\end{algorithmic}
\end{algorithm*}

\subsection{Prototypical alignment and pseudo label correction}
\label{III-D}
\subsubsection{Bidirectional prototype alignment}

Recent research~\citep{zhou2023semi} points out that the stochastic classifiers' parameters are equivalent to class prototypes, making it a highly efficient and elegant solution for prototype alignment.
By directly using these parameters as class prototypes, the classifier not only provides accurate prototype representations but also inherently aligns features with their corresponding prototypes during back propagation. Specifically, to ensure the stability of the model, we omit the standard deviation parameter in the stochastic classifiers. The classification decision using the learnable prototype cosine  similarity (CosSim) classifier is made as follows: 
\begin{equation}
\label{eqa7}
\begin{aligned}
    g^c(z) &= \operatorname{sim}(z, \mathbf{W}) / \tau_{temp}, \\
\end{aligned}
\end{equation}
where \( g^c \) denotes the CosSim classifier, and \( \operatorname{sim} \) denotes the cosine similarity function, \( z \in \mathbb{R}^{D'} \) is the feature vector, and \( \tau_{temp} \) is a temperature hyperparameter. \( \mathbf{W} = [w_1, ..., w_C] \), with each \( w_c \in \mathbb{R}^{D'} \) represents the trainable parameters corresponding to the class prototype for \( c = 1, ..., C \). Subsequently, we normalize the classifier output to the range \([0,1]\) to obtain the predicted probability map. Finally, we extend this formulation to the segmentation task, leading to the following equation: 
\begin{equation}
\label{eq:segmentation}
\begin{aligned}
    p^c = g^c(ft) = \operatorname{Softmax} \left(\operatorname{sim}(ft, \mathbf{W}) / \tau_{temp} \right), \\
\end{aligned}
\end{equation}
where \( ft \) denotes the segmentation feature map, and \( p^c \in \mathbb{R}^{C \times H \times W}\) represents the predicted probability map obtained using CosSim classifier, and we denote the predicted probability map from the Linear classifier as \( p^l \).

Based on the aforementioned concepts, to obtain two initially distinct but gradually converging CosSim classifier parameters for bidirectional prototype alignment, we first construct two classifiers, \( g^{c_1} \) and \( g^{c_2} \), with their respective parameters denoted as \( \mathbf{W}_1 \) and \( \mathbf{W}_2 \).  
Next, we define the parameters for the classifier \( g^{c_v} \), corresponding to the virtual domain, as \( \mathbf{W}_v = \frac{2\mathbf{W}_1 + (1+\lambda_{\text{sim}})\mathbf{W}_2}{3+\lambda_{\text{sim}}} \), while the parameters for the classifier \( g^{c_r} \), associated with the real domains, are given by \( \mathbf{W}_r = \frac{2\mathbf{W}_2 + (1+\lambda_{\text{sim}})\mathbf{W}_1}{3+\lambda_{\text{sim}}} \). The parameter \( \lambda_{\text{sim}} = e^{-5(1 - t/t_{\max})} \), which gradually increases towards 1 over training.

In our framework, there are two CosSim classifiers \( g^{c_v} \) and \( g^{c_r} \), as well as a Linear classifier \( g^{l} \). During the training process, we use \( g^{c_v} \) for training on the virtual domain, and \( g^{c_r} \) for training on real domains. The unique \( g^{l} \) is used across all data. Through this approach, bidirectional prototype alignment (BPA) enables smooth and steady prototype acquisition and feature alignment.

\subsubsection{Prototypical pseudo label correction}
\label{III-D2}
Linear and CosSim classifiers have distinct characteristics, and how to effectively utilize both to generate more accurate predictions is a significant research question. In our framework, the Linear classifier is trained on all data, resulting in a more stable linear decision boundary and relatively more accurate predictions. However, the training process inevitably generates discrete feature vectors that are linearly inseparable~\citep{zhang2021prototypical}, often leading to incorrect linear decisions and resulting in confirmation bias.

Considering the above mentioned considerations, we propose prototypical pseudo label correction (PPLC), which essentially uses \( p^c \) to correct \( p^l \). We first obtain the correction mask \( M_{\text{cr}} = \mathds{1}(\max(p^{c, \text{fore}}) > \tau) \), where \( p^{c, \text{fore}} \) is the predicted probability for the foreground classes, the indicator function is denoted as $\mathds{1}(\cdot)$, \( \tau \) is a pre-defined confidence threshold. When the value in \( M_{\text{cr}} \) is 1, it indicates that the prediction at the corresponding pixel in \( p^c \) is reliable and can be used to correct \( p^l \). Finally, we obtain the corrected predicted probability map \( p^{\text{cr}} \) as follows:  
\begin{equation}
\label{eqa10}
\begin{aligned}
    p^{\text{cr}} = p^c \odot M_{\text{cr}} + p^l \odot (1 - M_{\text{cr}}). 
\end{aligned}
\end{equation}

The aforementioned BPA and PPLC modules jointly constitute PAPLC component, which enables the acquisition of more discriminative feature representations and effectively alleviates confirmation bias.
\subsection{Loss function}
\label{III-E}
During the training process, the model is able to learn more efficiently from the unlabeled data in the virtual domain. The predictions $p^{w,v}$ generated by the teacher model on virtual unlabeled data may initially be more accurate than $p^{w}$. 
Therefore, we use the corrected predictions \( p^{cr,w,v} \) and \( p^{cr,w} \) to compute the average probability \( p^{w,avg} \) as the supervision signal for the student model:
\begin{equation}
\label{eqa11}
\begin{aligned}
p^{w,avg} = \frac{1}{2}(p^{cr,w,v} + p^{cr,w}),
\end{aligned}
\end{equation}
in particular, when the labeled and unlabeled data belong to the same domain, \( p^{w,avg} = p^{cr,w} \). Since the model utilizes the parameter averages of \( g^{c_1} \) and \( g^{c_2} \) as the final class prototypes, the teacher model using a CosSim classifier with the averaged parameters,  \( g^{c_{\text{avg}}} \) can generate more stable pseudo labels. Hence, for the teacher model, we use \( g^{c_{avg}} \), with the parameters given by:
\(\mathbf{W}_{avg} = \frac{\mathbf{W}_1 + \mathbf{W}_2}{2}
\).
To obtain more accurate supervision, we use a filtering mask \( m \) to indicate whether the pseudo labels are accurate:
\begin{equation}
\label{eqa12}
\begin{aligned}
m_i = \mathds{1} (\max(p^{w,avg}_i) > \tau),
\end{aligned}
\end{equation}
where \( m_i \) is \( i^{th} \) pixel of \( m \), 
the basic segmentation loss \( \mathcal{L}_{seg} = \mathcal{L}_{ce} + \mathcal{L}_{dice} \), which are formulated as:
\begin{equation}
\label{eqa13}
\begin{alignedat}{2}
    &\mathcal{L}_{seg}(y,p,m) &&= \mathcal{L}_{ce}(y,p,m) + \mathcal{L}_{dice}(y,p,m),\\
    &\mathcal{L}_{ce}(y,p,m) &&= -\frac{1}{H \times W}\sum_{i=1}^{H \times W} m_i y_i\log p_i, \\
    &\mathcal{L}_{dice}(y,p,m) &&= 1 - \frac{2 \times \sum_{i=1}^{H \times W} m_i p_i y_i}{\sum_{i=1}^{H \times W} m_i (p_i^2 + y_i^2)},
\end{alignedat}
\end{equation}
where \( p_i \), \( y_i \) denote the probability and pseudo label of the \( i^{th} \) pixel, respectively.

Given a pair of labeled data \( (p^\mathbf{1}, y_w) \) and unlabeled data \( (p^{w,avg}, \hat{p}^{w,avg}) \), \( p^{\mathbf{1}} \) represents the confidence of the ground truth, i.e., all are 1. According to Eq.~\eqref{eqa6}, we apply the same operation with \( M_1 \) and \( M_2 \) between \( p^\mathbf{1} \) and \( p^{w,avg} \), as well as between \( y_w \) and \( \hat{p}^{w,avg} \), which results in \(( p^{w,avg}_{in_{1/2}}, p^{w,avg}_{out_{1/2}}) \) and \(( \hat{p}^{w,avg}_{in_{1/2}}, \hat{p}^{w,avg}_{out_{1/2}}) \), the subscript \( 1/2 \) indicates the use of different \( M \). After the forward of the data in Eq.~\eqref{eqa6}, we obtain \( (p^{l/c_v,w,v}_{in}, p^{l/c_v,w,v}_{out}) \) and \( (p^{l/c_r,s}_{in}, p^{l/c_r,s}_{out}) \), with $l/c$ denoting the prediction from the Linear or CosSim classifier, and \(c_v/c_r \) corresponding to \(g^{c_r} / g^{c_v}\). Then, we can obtain the following losses:
\begin{equation}
\label{eqa14}
\begin{alignedat}{2}
    &\mathcal{L}_{seg,in_1} &&= \frac{1}{2}(\mathcal{L}^l_{seg,in_1}(\hat{p}^{w,avg}_{in_{1}}, p^{l,w,v}_{in}, m_{in_1}) + 
     \mathcal{L}^c_{seg,in_1}(\hat{p}^{w,avg}_{in_{1}}, p^{c_v,w,v}_{in}, m_{in_1})), \\
    &\mathcal{L}_{seg,out_1} &&= \frac{1}{2}(\mathcal{L}^l_{seg,out_1}(\hat{p}^{w,avg}_{out_{1}}, p^{l,w,v}_{out}, m_{out_1}) + 
  \mathcal{L}^c_{seg,out_1}(\hat{p}^{w,avg}_{out_{1}}, p^{c_v,w,v}_{out}, m_{out_1})),\\
    &\mathcal{L}_{seg,in_2} &&= \frac{1}{2}(\mathcal{L}^l_{seg,in_2}(\hat{p}^{w,avg}_{in_{2}}, p^{l,s}_{in}, m_{in_2}) + 
     \mathcal{L}^c_{seg,in_2}(\hat{p}^{w,avg}_{in_{2}}, p^{c_r,s}_{in}, m_{in_2})), \\
    &\mathcal{L}_{seg,out_2} &&= \frac{1}{2}(\mathcal{L}^l_{seg,out_2}(\hat{p}^{w,avg}_{out_{2}}, p^{l,s}_{out}, m_{out_2}) + 
     \mathcal{L}^c_{seg,out_2}(\hat{p}^{w,avg}_{out_{2}}, p^{c_r,s}_{out}, m_{out_2})),
\end{alignedat}
\end{equation}
where \( \mathcal{L}^l \) and \( \mathcal{L}^c \) represent the loss of the Linear and CosSim classifiers, respectively. Finally, the total loss of the proposed BCMDA framework is formulated as:
\begin{equation}
\label{eqa15}
\begin{aligned}
    \mathcal{L}_{total} = &\frac{1}{2}(\mathcal{L}_{seg,in_1} + \mathcal{L}_{seg,out_1}) + 
    \frac{1}{2}(\mathcal{L}_{seg,in_2} + \mathcal{L}_{seg,out_2}).
\end{aligned}
\end{equation}

\subsection{Inference}
\label{III-F}
During the inference phase, we adopt the student model for evaluation. As the decision outputs of \( g^{l,s} \) and \( g^{c_{\text{avg}},s} \) become increasingly consistent upon model convergence, we utilize only \( g^{l,s} \) at this stage. Specifically, for a test sample \( x^{\text{test}} \), the final prediction is obtained as \( \hat{p}^{\text{test}} = \arg\max\left(g^{l,s}(f^{s}(x^{\text{test}}))\right) \).

\begin{table}[!t]
\centering
\caption{Detailed information of all datasets.}
\vspace{+4mm}
\label{tab_ds}
\resizebox{1.0\columnwidth}{!}{
\begin{tabular}{cc|cccccc}
\toprule
Datasets                  & Configure        & Domain1  & Domain2  & Domain3 & Domain4 & Domain5 & Domain6 \\ \hline
\multirow{2}{*}{Fundus} & Name & Domain1  & Domain2  & Domain3  & Domain4  & \multirow{2}{*}{/} & \multirow{2}{*}{/} \\
                          & Train/Test Count & 50/51    & 99/60    & 320/80  & 320/80  &         &         \\ \hline
\multirow{2}{*}{Prostate} & Name             & RUNMC    & BMC      & HCRUDB  & UCL     & BIDMC   & HK      \\
                          & Train/Test Count & 338/83   & 305/79   & 373/95  & 133/42  & 225/36  & 136/22  \\ \hline
\multirow{2}{*}{M\&Ms}  & Name & Vendor A & Vendor B & Vendor C & Vendor D & \multirow{2}{*}{/} & \multirow{2}{*}{/} \\
                          & Train/Test Count & 1030/274 & 1342/325 & 525/129 & 550/135 &         &      \\     \bottomrule  
\end{tabular}
}
\end{table}

\section{Experiments}
\label{exper}
\subsection{Experimental setups}
\label{IV-A}

\subsubsection{Datasets}
\label{IV-A1} 
We evaluated our method on three public available multi-domain datasets: Fundus~\citep{wang2020dofe}, Prostate~\citep{liu2020shape}, and M\&Ms~\citep{campello2021multi}. In addition, we conducted experiments on the Synapse 3D dataset~\citep{landman2015miccai} for an abdominal multi-organ segmentation task under the SSMS setting. The Fundus dataset comprises retinal fundus images collected from four medical centers, used for the segmentation of the optic cup and optic disc. Each image was cropped to an 800 $\times$ 800 region of interest and subsequently resized and randomly cropped to 256 $\times$ 256 pixels using lanczos interpolation. The Prostate dataset consists of T2-weighted MRI images acquired from six different institutions for prostate segmentation, with each 2D slice resized to 384 $\times$ 384 pixels. 
The M\&Ms dataset was collected from four different MRI scanner vendors, with annotations available only for the end-systole and end-diastole phases, and involves the segmentation of three targets including the left ventricle (LV), left ventricular myocardium (MYO), and right ventricle (RV). 
Each slice was resized to 288 $\times$ 288 pixels. Following the SymGD~\citep{ma2024constructing} protocol, we adopted the same training and testing splits for each domain as in SymGD, with 789 and 271 2D slices for training and testing in the Fundus dataset, 1510 and 357 in the Prostate dataset, and 3447 and 863 in the M\&Ms dataset, respectively, as shown in Table~\ref{tab_ds}. 
All data were normalized to the range [-1, 1]. The Synapse dataset is a widely adopted benchmark for abdominal multi-organ segmentation and consists of 30 axial contrast-enhanced computed tomography (CT) volumes. Each volume is densely annotated with one background category and 13 foreground anatomical structures, including the spleen (Sp), right kidney (RK), left kidney (LK), gallbladder (Ga), esophagus (Es), liver (Li), stomach (St), aorta (Ao), inferior vena cava (IVC), portal and splenic veins (PSV), pancreas (Pa), right adrenal gland (RAG), and left adrenal gland (LAG). Following the experimental protocol of DHC~\citep{wang2023dhc}, the dataset was partitioned into 20 cases for training, 4 for validation, and 6 for testing.

\begin{table}[!t]
\centering
\caption{Comparison with other methods on the Fundus dataset, where \#L denotes the number of labeled samples. The best results are highlighted in \textbf{bold}, while the second-best results are \underline{underlined}.}
\vspace{+4mm}
\label{tab1}
\resizebox{1.0\columnwidth}{!}{
\begin{tabular}{l|c|llll|llll|c}
\hline
\multirow{2}{*}{Methods}             & \multicolumn{1}{l|}{\multirow{2}{*}{\#L}} & \multicolumn{4}{c|}{(Optic Cup / Optic Disc Segmentation) Dice $\uparrow$}                                                                        & Dice $\uparrow$               & Jaccard $\uparrow$            & 95HD $\downarrow$            & ASD $\downarrow$             & \multirow{2}{*}{$p$ value (Dice)} \\
                                     & \multicolumn{1}{l|}{}                     & \multicolumn{1}{c}{Domain 1}    & \multicolumn{1}{c}{Domain 2}    & \multicolumn{1}{c}{Domain 3}          & \multicolumn{1}{c|}{Domain 4}         & \multicolumn{4}{c|}{Avg.}                                                                                                   &                                   \\ \hline
U-Net                                & 20                                        & 57.92 / 71.80                   & 64.98 / 70.86                   & 48.88 / 65.03                         & 41.52 / 65.41                         & $60.80_{\pm1.17}$             & $51.99_{\pm0.93}$             & $47.62_{\pm0.93}$            & $29.28_{\pm0.60}$            & $<0.01$                           \\
Upper bound                          & *                                         & 85.47 / 93.38                   & 80.60 / 90.94                   & 85.28 / 93.02                         & 85.65 / 93.24                         & $88.45_{\pm0.02}$             & $80.32_{\pm0.04}$             & $7.42_{\pm0.01}$             & $3.73_{\pm0.04}$             & $<0.01$                           \\
UA-MT~\citep{yu2019uncertainty}      & 20                                        & 63.53 / 79.63                   & 67.60 / 79.69                   & 45.32 / 56.69                         & 41.63 / 60.45                         & $61.82_{\pm7.89}$             & $52.31_{\pm7.50}$             & $42.47_{\pm8.73}$            & $25.79_{\pm7.86}$            & $<0.01$                           \\
SIFA ~\citep{chen2020unsupervised}   & 20                                        & 59.89 / 75.98                   & 66.76 / 84.47                   & 64.47 / 85.31                         & 57.24 / 73.84                         & $70.99_{\pm4.54}$             & $58.48_{\pm5.25}$             & $18.59_{\pm2.22}$            & $11.76_{\pm1.17}$            & $<0.01$                           \\
FixMatch~\citep{sohn2020fixmatch}    & 20                                        & 81.32 / 91.49                   & 72.64 / 87.17                   & 78.72 / 92.78                         & 74.66 / 88.70                         & $83.44_{\pm0.06}$             & $73.76_{\pm0.39}$             & $11.35_{\pm0.59}$            & $5.61_{\pm0.01}$             & $<0.01$                           \\
UDA-VAE++~\citep{lu2022unsupervised} & 20                                        & 60.75 / 79.64                   & 69.16 / 86.38                   & 67.17 / 85.16                         & 68.63 / 79.56                         & $74.56_{\pm1.48}$             & $62.31_{\pm1.29}$             & $17.42_{\pm0.26}$            & $9.62_{\pm0.34}$             & $<0.01$                           \\
SS-Net~\citep{wu2022exploring}       & 20                                        & 61.25 / 79.90                   & 69.25 / 83.73                   & 53.88 / 69.43                         & 39.56 / 62.54                         & $64.94_{\pm2.50}$             & $55.36_{\pm2.65}$             & $40.56_{\pm6.14}$            & $23.62_{\pm2.98}$            & $<0.01$                           \\
BCP~\citep{bai2023bidirectional}     & 20                                        & 75.29 / 90.63                   & 78.92 / \underline{91.73}       & 76.63 / 90.54                         & 77.81 / 90.78                         & $84.04_{\pm1.40}$             & $74.78_{\pm1.58}$             & 10.52$_{\pm0.75}$            & $5.40_{\pm0.56}$             & $<0.01$                           \\
CauSSL~\citep{miao2023caussl}        & 20                                        & 66.53 / 83.38                   & 67.28 / 81.66                   & 55.36 / 71.80                         & 39.36 / 48.04                         & $64.17_{\pm3.34}$             & $54.15_{\pm3.32}$             & $36.44_{\pm6.80}$            & $21.85_{\pm2.96}$            & $<0.01$                           \\
ABD~\citep{chi2024adaptive}          & 20                                        & 76.72 / 86.04                   & 77.47 / 90.56                   & 76.78 / 87.27                         & 74.78 / 88.72                         & $82.29_{\pm0.36}$             & $72.55_{\pm0.34}$             & $12.27_{\pm0.66}$            & $6.76_{\pm0.29}$             & $<0.01$                           \\
SDCL~\citep{song2024sdcl}            & 20                                        & 80.59 / 91.65                   & \underline{79.73} / 90.62       & 79.28 / 90.61                         & 80.66 / 91.46                         & $85.58_{\pm0.19}$             & $76.73_{\pm0.37}$             & $9.58_{\pm0.22}$             & $5.10_{\pm0.15}$             & $<0.01$                           \\
SymGD~\citep{ma2024constructing}     & 20                                        & \underline{83.16} / \underline{92.91}       & 79.03 / 89.49                   & \underline{83.28} / \underline{92.82} & \underline{83.38} / \underline{93.50} & $\underline{87.20}_{\pm0.66}$ & $\underline{78.46}_{\pm0.91}$ & $\underline{8.09}_{\pm0.18}$ & $\underline{3.92}_{\pm0.04}$ & $<0.01$                           \\
\textbf{BCMDA (ours)}                & 20                                        & \textbf{84.25} / \textbf{93.68} & \textbf{83.35} / \textbf{92.46} & \textbf{84.67} / \textbf{93.51}       & \textbf{85.53} / \textbf{94.12}       & $\textbf{88.95}_{\pm0.15}$    & $\textbf{81.03}_{\pm0.22}$    & $\textbf{7.21}_{\pm0.11}$    & $\textbf{3.45}_{\pm0.03}$    & -                                 \\ \hline
\end{tabular}
}
\end{table}
\begin{figure}[!t]
\centerline{\includegraphics[width=0.99\columnwidth]{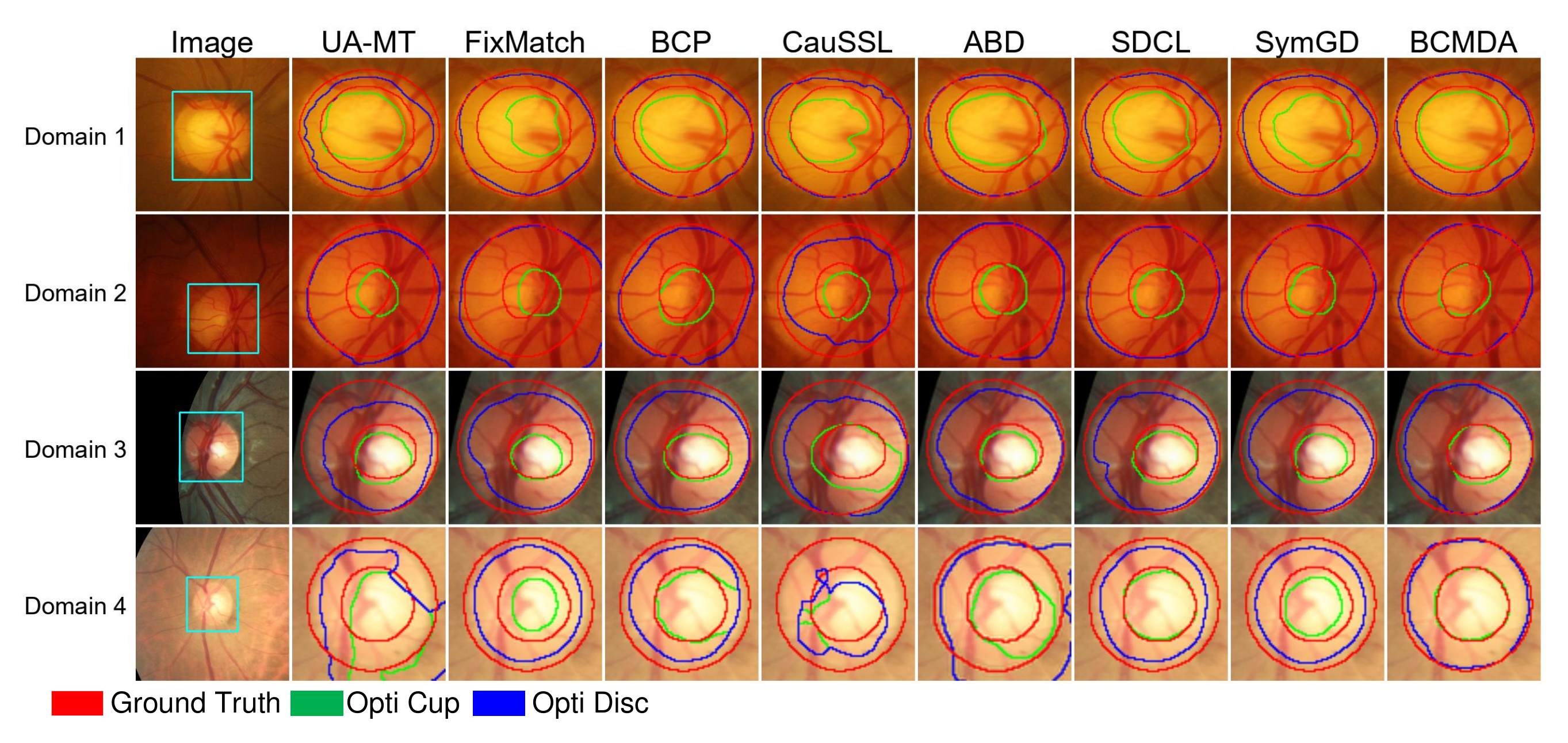}}
\vspace{-4mm}
\caption{Visualization results on the Fundus dataset.}
\label{fundus}
\end{figure}

\subsubsection{Implementation details}
\label{IV-A2}
Our method was implemented in Python based on the PyTorch framework and trained using an NVIDIA GeForce RTX 4090 GPU. We adopted U-Net~\citep{ronneberger2015u} as the backbone architecture and set the $\tau = 0.95$. For the Fundus dataset, we set $\lambda_{\text{fix}}$ to 0.75, $\alpha$ to 0.70, and $\tau_{\text{temp}}$ to 0.05, while for the Prostate and M\&Ms datasets, $\lambda_{\text{fix}}$ was set to 0.65, $\alpha$ to 1.0, and $\tau_{\text{temp}}$ to 0.05 and 0.30, respectively. We also set $W' = \frac{1}{4} W$. Following the SymGD~\citep{ma2024constructing}, we used Stochastic Gradient Descent (SGD) with an initial learning rate of 0.03, a momentum of 0.9, and a weight decay of 0.0001.
The batch size was fixed at 8, equally divided between labeled and unlabeled data. The number of training iterations was set to 30,000 for the Fundus dataset and 60,000 for both the Prostate and M\&Ms datasets. 
Specifically, for tasks in the M\&Ms dataset involving only 5 labeled samples, the batch size was reduced to 4. In each experiment, we designated one domain as the source of labeled data, while the remaining data across all domains were used as unlabeled data. During evaluation, the performance on each domain was assessed individually, and the final result was reported as the average performance across all domains, as shown in the Domain1 column of Table~\ref{tab1}. 
For the Synapse dataset, $\lambda_{\text{fix}}$ was set to 0.65, $\alpha$ was set to 1.0, and $\tau_{\text{temp}}$ was set to 0.05. The number of iterations was 3000, the batch size was 4, and V-Net~\citep{milletari2016v} was used as the backbone. During testing, segmentation predictions were produced using a sliding-window inference strategy with a stride of 32 $\times$ 32 $\times$ 16. We employed four evaluation metrics: Dice, Jaccard, Average Surface Distance (ASD), and 95\% Hausdorff Distance (95HD).
\begin{table}[!t]
\centering
\caption{Comparison with other methods on the Prostate dataset.}
\vspace{+4mm}
\label{tab2}
\resizebox{1.0\columnwidth}{!}{
\begin{tabular}{l|c|cccccc|cccc|c}
\hline
\multirow{2}{*}{Methods}          & \multirow{2}{*}{\#L} & \multicolumn{6}{c|}{(Prostate Segmentation) Dice $\uparrow$}                                                          & Dice $\uparrow$               & Jaccard $\uparrow$            & 95HD $\downarrow$             & ASD $\downarrow$              & \multirow{2}{*}{$p$ value (Dice)} \\
                                  &                      & RUNMC             & BMC               & HCRUDB            & UCL               & BIDMC             & HK                & \multicolumn{4}{c|}{Avg.}                                                                                                     &                                   \\ \hline
U-Net                             & 20                   & 20.79             & 19.02             & 17.74             & 15.25             & 18.91             & 29.12             & $20.14_{\pm0.45}$             & $14.82_{\pm1.30}$             & $121.66_{\pm6.20}$            & $82.41_{\pm3.10}$             & $<0.01$                           \\
U-Net                             & 40                   & 28.71             & 35.61             & 21.23             & 43.56             & 20.60             & 27.62             & $29.56_{\pm1.62}$             & $24.03_{\pm1.12}$             & $97.97_{\pm4.04}$             & $65.18_{\pm0.94}$             & $<0.01$                           \\
Upper bound                       & *                    & 88.56             & 88.56             & 85.60             & 88.63             & 88.90             & 89.44             & $88.28_{\pm0.05}$             & $80.67_{\pm0.05}$             & $10.07_{\pm0.04}$             & $4.13_{\pm0.02}$              & $<0.01$                           \\
UA-MT~\citep{yu2019uncertainty}   & 20                   & 18.44             & 15.17             & 16.15             & 33.54             & 14.32             & 13.50             & $18.52_{\pm0.55}$             & $13.21_{\pm0.34}$             & $129.30_{\pm2.26}$            & $87.65_{\pm2.80}$             & $<0.01$                           \\
FixMatch~\citep{sohn2020fixmatch} & 20                   & 82.93             & 64.24             & 54.11             & 70.23             & 11.15             & 82.82             & $60.91_{\pm0.89}$             & $53.22_{\pm0.73}$             & $46.61_{\pm4.02}$             & $26.85_{\pm2.13}$             & $<0.01$                           \\
SS-Net~\citep{wu2022exploring}    & 20                   & 15.28             & 12.25             & 15.46             & 35.63             & 16.29             & 13.21             & $18.02_{\pm1.51}$             & $13.90_{\pm2.07}$             & $115.52_{\pm7.19}$            & $78.77_{\pm3.98}$             & $<0.01$                           \\
BCP~\citep{bai2023bidirectional}  & 20                   & 61.26             & 62.53             & 51.82             & 55.91             & 61.38             & 62.19             & $59.18_{\pm0.48}$             & $50.64_{\pm2.85}$             & $55.46_{\pm1.47}$             & $26.49_{\pm1.54}$             & $<0.01$                           \\
CauSSL~\citep{miao2023caussl}     & 20                   & 22.75             & 33.42             & 17.57             & 29.72             & 28.26             & 28.73             & $26.74_{\pm3.97}$             & $22.39_{\pm5.34}$             & $112.85_{\pm3.63}$            & $66.26_{\pm5.82}$             & $<0.01$                           \\
ABD~\citep{chi2024adaptive}       & 20                   & 55.82             & 64.39             & 11.14             & 61.37             & 53.84             & 24.28             & $45.14_{\pm2.56}$             & $34.53_{\pm3.29}$             & $74.58_{\pm1.74}$             & $45.83_{\pm2.01}$             & $<0.01$                           \\
SDCL~\citep{song2024sdcl}         & 20                   & 75.91             & \underline{77.18} & 72.29             & 80.86             & \underline{70.17} & 68.95             & $74.23_{\pm0.32}$             & $64.39_{\pm0.25}$             & $\underline{22.63}_{\pm1.91}$ & $\underline{11.78}_{\pm1.08}$ & $<0.01$                           \\
SymGD~\citep{ma2024constructing}  & 20                   & \underline{88.36} & 73.66             & \underline{76.44} & \underline{84.28} & 51.39             & \underline{81.20} & $\underline{75.89}_{\pm1.36}$ & $\underline{66.45}_{\pm2.03}$ & $41.91_{\pm4.09}$             & $21.95_{\pm1.22}$             & $<0.01$                           \\
\textbf{BCMDA (ours)}             & 20                   & \textbf{88.54}    & \textbf{84.02}    & \textbf{86.75}    & \textbf{87.03}    & \textbf{86.15}    & \textbf{88.81}    & $\textbf{86.88}_{\pm0.15}$    & $\textbf{78.47}_{\pm0.42}$    & $\textbf{12.04}_{\pm0.26}$    & $\textbf{5.12}_{\pm0.27}$     & -                                 \\ \hline
U-Net                             & 20                   & 20.79             & 19.02             & 17.74             & 15.25             & 18.91             & 29.12             & $20.14_{\pm0.45}$             & $14.82_{\pm1.30}$             & $121.66_{\pm6.20}$            & $82.41_{\pm3.10}$             & $<0.01$                           \\
U-Net                             & 40                   & 28.71             & 35.61             & 21.23             & 43.56             & 20.60             & 27.62             & $29.56_{\pm1.62}$             & $24.03_{\pm1.12}$             & $97.97_{\pm4.04}$             & $65.18_{\pm0.94}$             & $<0.01$                           \\
Upper bound                       & *                    & 88.56             & 88.56             & 85.60             & 88.63             & 88.90             & 89.44             & $88.28_{\pm0.05}$             & $80.67_{\pm0.05}$             & $10.07_{\pm0.04}$             & $4.13_{\pm0.02}$              & 0.13                              \\
UA-MT~\citep{yu2019uncertainty}   & 40                   & 27.41             & 16.66             & 9.91              & 40.40             & 15.57             & 14.65             & $20.77_{\pm0.57}$             & $15.00_{\pm0.17}$             & $121.23_{\pm12.95}$           & $79.17_{\pm2.26}$             & $<0.01$                           \\
FixMatch~\citep{sohn2020fixmatch} & 40                   & 80.59             & 71.06             & 73.16             & 77.12             & 60.54             & 84.95             & $74.57_{\pm0.23}$             & $65.78_{\pm0.26}$             & $23.11_{\pm1.51}$             & $13.02_{\pm1.51}$             & $<0.01$                           \\
SS-Net~\citep{wu2022exploring}    & 40                   & 25.58             & 24.98             & 14.61             & 43.67             & 18.43             & 9.57              & $22.80_{\pm2.11}$             & $17.25_{\pm2.11}$             & $109.57_{\pm0.04}$            & $72.94_{\pm2.56}$             & $<0.01$                           \\
BCP~\citep{bai2023bidirectional}  & 40                   & 65.39             & 73.27             & 47.41             & 63.64             & 74.75             & 62.62             & $64.51_{\pm0.42}$             & $54.65_{\pm0.73}$             & $51.07_{\pm2.16}$             & $25.46_{\pm2.49}$             & $<0.01$                           \\
CauSSL~\citep{miao2023caussl}     & 40                   & 26.23             & 37.11             & 17.76             & 35.45             & 24.99             & 24.63             & $27.70_{\pm9.57}$             & $21.12_{\pm7.97}$             & $111.58_{\pm4.30}$            & $67.71_{\pm7.90}$             & $<0.01$                           \\
ABD~\citep{chi2024adaptive}       & 40                   & 60.94             & 68.70             & 13.98             & 58.22             & 72.07             & 24.77             & $49.78_{\pm0.11}$             & $40.44_{\pm0.05}$             & $64.46_{\pm5.64}$             & $37.38_{\pm2.45}$             & $<0.01$                           \\
SDCL~\citep{song2024sdcl}         & 40                   & 76.55             & 79.24             & 77.40             & 81.13             & 81.32             & 72.67             & $78.05_{\pm0.20}$             & $69.05_{\pm0.15}$             & $19.08_{\pm0.86}$             & $10.61_{\pm1.33}$             & $<0.01$                           \\
SymGD~\citep{ma2024constructing}  & 40                   & \underline{86.80} & \underline{86.34} & \underline{84.88} & \underline{86.69} & \textbf{88.36} & \underline{87.09} & $\underline{86.69}_{\pm1.82}$ & $\underline{78.75}_{\pm2.07}$ & $\underline{11.83}_{\pm2.07}$ & $\underline{4.98}_{\pm1.11}$  & $<0.01$                           \\
\textbf{BCMDA (ours)}             & 40                   & \textbf{89.06}    & \textbf{86.80}    & \textbf{88.71}    & \textbf{88.35}    & \underline{87.97}    & \textbf{89.24}    & $\textbf{88.36}_{\pm0.28}$    & $\textbf{80.58}_{\pm0.30}$    & $\textbf{10.42}_{\pm0.39}$    & $\textbf{4.09}_{\pm0.36}$     & -                                 \\ \hline
\end{tabular}
}
\end{table}
\begin{figure}[!t]
\centerline{\includegraphics[width=0.99\columnwidth]{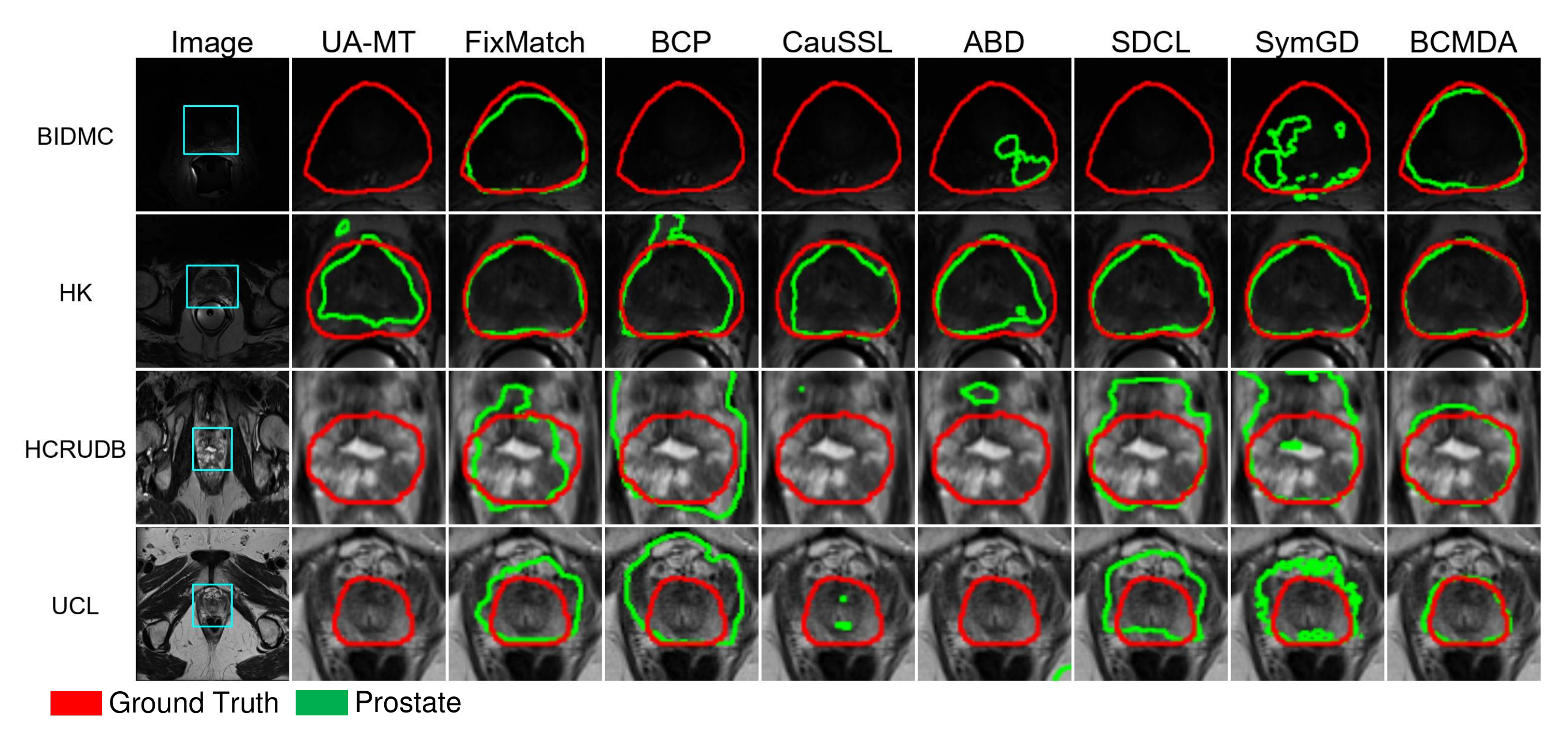}}
\vspace{-4mm}
\caption{Visualization results on the Prostate dataset using 20 labeled data.}
\label{prostate}
\end{figure}

\subsection{Comparison with other methods}
\label{IV-B}
We compared BCMDA with current SOTA SSMS and UDA methods. Specifically, we evaluated UA-MT~\citep{yu2019uncertainty}, FixMatch~\citep{sohn2020fixmatch}, SS-Net~\citep{wu2022exploring}, BCP~\citep{bai2023bidirectional}, CauSSL~\citep{miao2023caussl}, ABD~\citep{chi2024adaptive}, SDCL~\citep{song2024sdcl}, and SymGD~\citep{ma2024constructing} across three datasets. For the Fundus dataset, we additionally compared SIFA~\citep{chen2020unsupervised} and UDA-VAE++~\citep{lu2022unsupervised}. According to SymGD~\citep{ma2024constructing}, an upper bound was obtained by incorporating UCP into FixMatch and using all data from the corresponding domain for the labeled samples. The U-Net represents the baseline trained solely on labeled data. Furthermore, on the Synapse dataset, we evaluated several SOTA methods, including URPC~\citep{luo2022semiurpc}, CPS~\citep{chen2021cps}, CLD~\citep{lin2022calibrating}, DHC~\citep{wang2023dhc}, AllSpark~\citep{wang2024allspark}, GenericSSL~\citep{wang2023towards}, and S\&D-Messenger~\citep{zhang2025s}.

\subsubsection{Fundus dataset}
\label{IV-B1}
Table~\ref{tab1} presents the performance of all methods under different labeled domains using 20 labeled samples, along with the average performance. The proposed approach achieves best performance across all evaluated metrics, with average Dice of 88.95\% and Jaccard of 81.03\%, outperforming previous SOTA method SymGD by 1.75\% in Dice. It also demonstrates strong robustness, with low standard deviations of 0.15 for Dice and 0.22 for Jaccard. In terms of geometric accuracy, the proposed approach yields lowest boundary errors, achieving 95HD of 7.21\,mm and ASD of 3.45\,mm with low standard deviations. Moreover, consistent improvements are observed across all metrics, even when compared with the upper bound, with a 0.5\% gain in Dice. Statistical significance tests confirm the effectiveness of the proposed approach, with all $p$-values for Dice comparisons against other methods below 0.01. Fig.~\ref{fundus} presents a qualitative comparison of optic cup and optic disc segmentation results across several SOTA approaches, with samples selected from different domains. Although SymGD achieves competitive performance compared to earlier methods such as FixMatch and BCP, it still suffers from boundary ambiguity and reduced topological smoothness under complex retinal vessel occlusions. In contrast, our method produces segmentation boundaries that more closely align with the ground-truth contours, yielding more accurate and robust delineations of both the optic cup and disc, particularly on challenging domain 1, where SymGD exhibits slight over-segmentation, thereby highlighting the advantage of our method. 
The left panel of Fig.~\ref{visualcom} compares our method with SOTA approaches, including FixMatch, SDCL, and SymGD, on the Fundus dataset. Our method consistently achieves the highest Dice scores across all four domains. These results further demonstrate the superiority of the proposed method.

\subsubsection{Prostate dataset}
\label{IV-B2}
We present experimental results on the Prostate dataset using 40 and 20 labeled samples. As shown in Table~\ref{tab2}, our method achieves the highest scores across all average metrics. With 40 labeled samples, it attains average Dice of 88.36\% and ASD of 4.09\,mm, even surpassing the upper bound performance of 88.28\% and 4.13\,mm. When the number of labeled samples is reduced to 20, the proposed approach still maintains strong performance, achieving average Dice of 86.88\%, which represents substantial improvement of 10.99\% over the previous SOTA method SymGD with Dice of 75.89\%. Moreover, superior stability is observed, as indicated by remarkably low standard deviations of 0.15 for Dice and 0.27 for ASD, whereas SymGD exhibits much higher standard deviations of 1.36 and 1.22, respectively.
\begin{table}[!t]
\centering
\caption{
Comparison with other methods on the M\&Ms dataset.
}
\vspace{+4mm}
\label{tab3}
\resizebox{\columnwidth}{!}{
  \begin{tabular}{l|c|cccc|cccc|c}
\hline
\multirow{2}{*}{Method}           & \multirow{2}{*}{\#L} & \multicolumn{4}{c|}{(LV / MYO / RV Segmentation) Dice $\uparrow$}                                                                                                                                                                    & Dice $\uparrow$                & Jaccard $\uparrow$            & 95HD $\downarrow$            & ASD $\downarrow$             & \multirow{2}{*}{$p$ value (Dice)} \\
                                  &                      & Vendor A                                            & Vendor B                                               & Vendor C                                                  & Vendor D                                                  & \multicolumn{4}{c|}{Avg.}                                                                                                    &                                   \\ \hline
U-Net                             & 5                    & 33.87 / 17.97 / 27.00                               & 48.14 / 49.23 / 29.02                                  & 16.22 / 13.37 / 19.18                                     & 43.20 / 34.06 / 35.37                                     & $30.56_{\pm1.28}$              & $22.19_{\pm3.51}$             & $68.44_{\pm3.64}$            & $44.32_{\pm6.02}$            & $<0.01$                           \\
U-Net                             & 20                   & 41.67 / 27.89 / 28.39                               & 56.56 / 49.59 / 47.19                                  & 55.89 / 45.64 / 42.60                                     & 69.31 / 58.30 / 57.52                                     & $48.38_{\pm6.55}$              & $40.61_{\pm5.22}$             & $43.02_{\pm7.00}$            & $28.68_{\pm8.20}$            & $<0.01$                           \\
Upper bound                       & *                    & 91.19 / 83.65 / 84.00                               & 92.48 / 85.56 / 84.54                                  & 92.26 / 84.35 / 84.27                                     & 91.82 / 82.90 / 83.50                                     & $86.71_{\pm0.15}$              & $78.52_{\pm0.17}$             & $4.41_{\pm0.13}$             & $2.24_{\pm0.13}$             & $<0.01$                           \\
UA-MT~\citep{yu2019uncertainty}   & 5                    & 16.32 / 8.85 / 10.54                                & 38.10 / 36.26 / 16.47                                  & 19.29 / 13.89 / 12.13                                     & 33.26 / 21.99 / 16.00                                     & $20.26_{\pm0.51}$              & $14.73_{\pm0.59}$             & $77.47_{\pm1.17}$            & $54.19_{\pm0.82}$            & $<0.01$                           \\
FixMatch~\citep{sohn2020fixmatch} & 5                    & \underline{80.67} / 66.81 / 57.78                   & 85.09 / 80.19 / 75.35                                  & 83.18 / \underline{75.71} / 67.40                         & \underline{89.60} / \underline{78.44} / \underline{79.04} & $76.61_{\pm1.69}$              & $66.85_{\pm0.81}$             & $\underline{9.65}_{\pm0.80}$             & $5.12_{\pm0.36}$             & $<0.01$                           \\
SS-Net~\citep{wu2022exploring}    & 5                    & 12.94 / 11.15 / 3.98                                & 31.74 / 28.30 / 22.10                                  & 6.86 / 6.46 / 5.68                                        & 23.98 / 16.06 / 4.84                                      & $14.51_{\pm1.13}$              & $12.90_{\pm3.34}$             & $83.77_{\pm0.84}$            & $62.99_{\pm1.62}$            & $<0.01$                           \\
BCP~\citep{bai2023bidirectional}  & 5                    & 73.18 / 32.47 / 37.45                               & 67.19 / 55.86 / 57.55                                  & 64.41 / 40.52 / 43.89                                     & 66.91 / 62.04 / 59.08                                     & $55.04_{\pm2.81}$              & $45.63_{\pm1.74}$             & $35.74_{\pm2.09}$            & $23.48_{\pm1.08}$            & $<0.01$                           \\
CauSSL~\citep{miao2023caussl}     & 5                    & 43.82 / 26.77 / 19.79                               & 47.68 / 37.32 / 21.71                                  & 9.86 / 16.53 / 16.22                                      & 32.68 / 30.75 / 29.86                                     & $27.75_{\pm4.23}$              & $20.12_{\pm3.35}$             & $64.66_{\pm5.99}$            & $38.17_{\pm3.72}$            & $<0.01$                           \\
ABD~\citep{chi2024adaptive}       & 5                    & 41.40 / 24.88 / 5.94                                & 25.31 / 21.25 / 16.72                                  & 17.70 / 9.43 / 11.57                                      & 34.98 / 29.56 / 30.42                                     & $22.43_{\pm0.76}$              & $17.04_{\pm0.30}$             & $63.68_{\pm0.46}$            & $52.05_{\pm0.98}$            & $<0.01$                           \\
SDCL~\citep{song2024sdcl}         & 5                    & 77.44 / 61.02 / 69.62                               & 88.43 / 77.91 / 76.79                                  & 81.85 / 69.38 / \underline{71.61}                         & 84.75 / 76.03 / 81.62                                     & $76.37_{\pm0.43}$              & $66.53_{\pm0.35}$             & $12.08_{\pm0.22}$            & $7.72_{\pm0.34}$             & $<0.01$                           \\
SymGD~\citep{ma2024constructing}  & 5                    & 64.00 / \underline{74.02} / \textbf{72.90}          & \underline{88.82} / \underline{80.23} / \textbf{82.98} & \underline{83.85} / 72.09 / 71.53                         & 86.80 / 76.69 / 69.39                                     & $\underline{77.78}_{\pm1.30}$  & $\underline{67.40}_{\pm1.46}$ & $13.35_{\pm1.04}$ & $\underline{5.26}_{\pm1.40}$ & $<0.01$                           \\
\textbf{BCMDA (ours)}             & 5                    & \textbf{83.36} / \textbf{75.97} / \underline{71.78} & \textbf{89.00} / \textbf{80.84} / \underline{82.08}    & \textbf{86.22} / \textbf{79.63} / \textbf{81.75}       & \textbf{91.32} / \textbf{82.81} / \textbf{82.83}          & $\textbf{82.30}_{\pm0.16}$     & $\textbf{72.53}_{\pm0.19}$    & $\textbf{6.46}_{\pm0.20}$    & $\textbf{2.97}_{\pm0.11}$    & -                                 \\ \hline
U-Net                             & 5                    & 33.87 / 17.97 / 27.00                               & 48.14 / 49.23 / 29.02                                  & 16.22 / 13.37 / 19.18                                     & 43.20 / 34.06 / 35.37                                     & $30.56_{\pm1.28}$              & $22.19_{\pm3.51}$             & $68.44_{\pm3.14}$            & $44.32_{\pm6.02}$            & $<0.01$                           \\
U-Net                             & 20                   & 41.67 / 27.89 / 28.39                               & 56.56 / 49.59 / 47.19                                  & 55.89 / 45.64 / 42.60                                     & 69.31 / 58.30 / 57.52                                     & $48.38_{\pm6.55}$              & $40.61_{\pm5.22}$             & $43.02_{\pm7.00}$            & $28.68_{\pm8.20}$            & $<0.01$                           \\
Upper bound                       & *                    & 91.19 / 83.65 / 84.00                               & 92.48 / 85.56 / 84.54                                  & 92.26 / 84.35 / 84.27                                     & 91.82 / 82.90 / 83.50                                     & $86.71_{\pm0.15}$              & $78.52_{\pm0.17}$             & $4.41_{\pm0.13}$             & $2.24_{\pm0.13}$             & $<0.01$                           \\
UA-MT~\citep{yu2019uncertainty}   & 20                   & 40.48 / 30.44 / 20.91                               & 50.65 / 43.24 / 36.96                                  & 48.55 / 37.35 / 30.98                                     & 47.27 / 43.11 / 34.01                                     & $38.66_{\pm0.99}$              & $30.14_{\pm1.42}$             & $71.75_{\pm0.85}$            & $41.14_{\pm0.43}$            & $<0.01$                           \\
FixMatch~\citep{sohn2020fixmatch} & 20                   & 87.97 / \underline{78.70} / 78.95                   & 90.32 / 81.61 / 80.80                                  & 88.25 / 80.37 / 75.56                                     & 90.42 / 81.20 / 81.48                                     & $82.97_{\pm0.02}$              & $74.02_{\pm0.04}$             & $6.49_{\pm0.39}$             & $3.45_{\pm0.09}$             & $<0.01$                           \\
SS-Net~\citep{wu2022exploring}    & 20                   & 40.10 / 25.92 / 21.34                               & 49.29 / 44.00 / 39.09                                  & 58.61 / 48.42 / 45.58                                     & 51.87 / 48.06 / 36.84                                     & $42.43_{\pm2.32}$              & $34.88_{\pm1.43}$             & $53.40_{\pm5.51}$            & $36.52_{\pm5.62}$            & $<0.01$                           \\
BCP~\citep{bai2023bidirectional}  & 20                   & 85.53 / 72.82 / 76.50                               & 72.51 / 63.16 / 64.75                                  & 71.35 / 61.58 / 62.13                                     & 83.45 / 69.48 / 81.10                                     & $72.03_{\pm0.54}$              & $62.75_{\pm0.12}$             & $28.23_{\pm3.79}$            & $15.70_{\pm3.57}$            & $<0.01$                           \\
CauSSL~\citep{miao2023caussl}     & 20                   & 43.06 / 28.50 / 16.39                               & 47.72 / 38.45 / 30.91                                  & 48.55 / 38.76 / 30.49                                     & 52.30 / 44.14 / 34.88                                     & $37.84_{\pm3.40}$              & $28.82_{\pm2.95}$             & $75.07_{\pm3.07}$            & $38.74_{\pm1.06}$            & $<0.01$                           \\
ABD~\citep{chi2024adaptive}       & 20                   & 43.08 / 36.34 / 28.46                               & 46.57 / 28.19 / 28.62                                  & 63.78 / 43.19 / 47.49                                     & 66.59 / 53.55 / 43.67                                     & $44.13_{\pm0.47}$              & $35.67_{\pm0.73}$             & $51.84_{\pm0.58}$            & $35.69_{\pm1.16}$            & $<0.01$                           \\
SDCL~\citep{song2024sdcl}         & 20                   & 87.21 / 76.16 / 78.82                               & 89.48 / \underline{82.91} / 72.14                      & 87.32 / 80.02 / 81.21                                     & 88.50 / 81.03 / 82.76                                     & $82.30_{\pm0.68}$              & $73.29_{\pm0.32}$             & $6.84_{\pm0.41}$             & $3.91_{\pm0.25}$             & $<0.01$                           \\
SymGD~\citep{ma2024constructing}  & 20                   & \underline{88.36} / 78.28 / \underline{81.29}       & \underline{90.36} / 81.78 / \underline{82.58}          & \underline{90.13} / \underline{82.75} / \underline{82.19} & \underline{91.59} / \underline{82.47} / \underline{84.15} & $\underline{84.66}_{\pm.0.50}$ & $\underline{75.69}_{\pm0.73}$ & $\underline{4.89}_{\pm0.37}$ & $\underline{2.37}_{\pm0.07}$ & $<0.01$                           \\
\textbf{BCMDA (ours)}             & 20                   & \textbf{88.95} / \textbf{79.63} / \textbf{82.01}    & \textbf{92.06} / \textbf{84.98} / \textbf{84.25}    & \textbf{90.61} / \textbf{83.48} / \textbf{84.03}          & \textbf{92.00} / \textbf{83.17} / \textbf{84.23}          & $\textbf{85.78}_{\pm.0.23}$    & $\textbf{76.81}_{\pm.0.26}$   & $\textbf{4.12}_{\pm0.19}$    & $\textbf{1.75}_{\pm0.14}$    & -                                 \\ \hline
\end{tabular}
}
\end{table}
\begin{figure}[!t]
\centerline{\includegraphics[width=0.99\columnwidth]{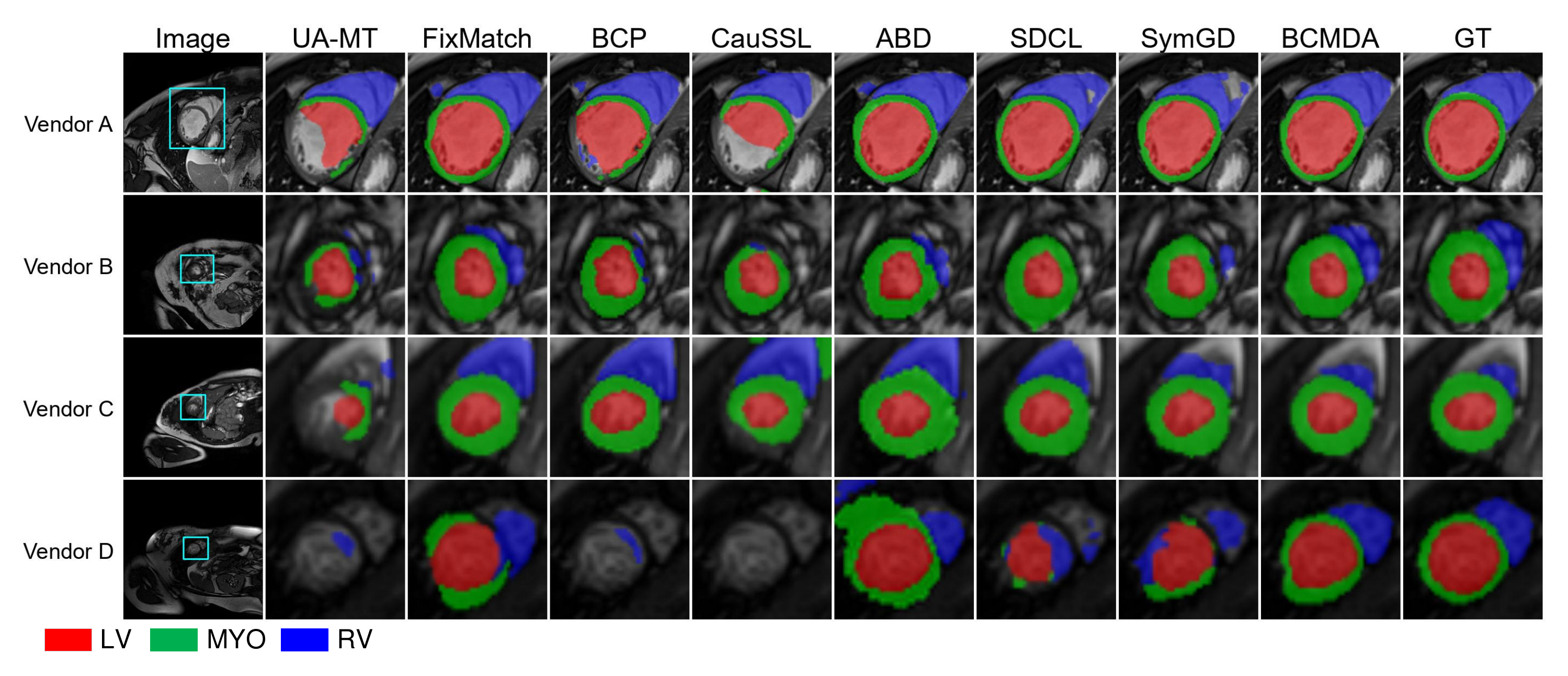}}
\vspace{-4mm}
\caption{Visualization results on the M\&Ms dataset using 20 labeled data.}
\label{MNMS}
\end{figure}These results demonstrate that our method exhibits minimal performance variation as the amount of labeled data decreases, with the Dice dropping only slightly from 88.36\% to 86.88\% when the labeled set is halved, while maintaining statistically significant improvements with $p$-values below 0.01. Challenging cases from different domains are visualized in Fig.~\ref{prostate}. ABD and SDCL exhibit noticeable morphological deviations, often failing to suppress false positives or accurately delineate the prostate structure in the presence of complex background noise and interference from small surrounding tissues. Even the previous SOTA method SymGD frequently struggles with blurred boundaries, leading to fragmented predictions or over-segmentation. In contrast, our method consistently produces smooth and coherent contours that closely align with the red ground truth, even in challenging domains such as HCRUDB and UCL, where low contrast and significant interference from adjacent tissues are prevalent. This demonstrates that the proposed approach can effectively handle ambiguous boundary information and deliver robust and reliable segmentation results. In addition, the right panel of Fig.~\ref{visualcom} illustrates the performance advantage of our method over other approaches, revealing a substantial improvement.

\subsubsection{M\&Ms dataset}
\label{IV-B3}
In Table~\ref{tab3}, the experimental results on the M\&Ms dataset using 20 and 5 labeled samples are presented. With 20 labeled samples, the proposed approach achieves performance comparable to or even exceeding the upper bound, particularly in terms of geometric metrics, obtaining 95HD of 4.12\,mm and ASD of 1.75\,mm. Meanwhile, the standard deviations are only 0.23 for Dice and 0.14 for ASD. When the number of labeled samples is reduced to 5, our method still maintains strong performance, achieving Dice of 82.30\%, with only minimal gap to the upper bound and a clear advantage over other frameworks such as FixMatch and SDCL. Even when compared with SymGD, the proposed approach demonstrates a substantial 4.52\% improvement. All improvements are statistically significant, as indicated by $p$-values below 0.01. The segmentation visualizations in Fig.~\ref{MNMS} further demonstrate that our method achieves the highest structural integrity among all compared approaches. Across diverse imaging data from Vendors A, B, C, and D, frameworks such as UA-MT and BCP often produce fragmented results or fail to preserve the correct topological relationships among cardiac components, particularly in the challenging Vendor D case where image contrast is limited. In addition, in a representative sample from Vendor B, other methods struggle to correctly segment the right ventricle (RV), whereas our method can accurately identify this structure. While SymGD frequently misidentifies boundary pixels or generates irregular shapes, our approach consistently yields smooth and anatomically plausible segmentations that closely align with the ground truth (GT). Overall, these results confirm that the proposed method can effectively segment complex cardiac structures across diverse domains, providing reliable and robust performance.

\begin{table}[!t]
\centering
\caption{
Comparison with other methods on the Synapse dataset.
}
\vspace{+4mm}
\label{tabsyn}
\resizebox{\columnwidth}{!}{
\begin{tabular}{l|cc|c|ccccccccccccc}
\hline
\multirow{2}{*}{Methods}           & \multirow{2}{*}{Avg. Dice $\uparrow$} & \multirow{2}{*}{Avg. ASD $\downarrow$} & \multicolumn{1}{l|}{\multirow{2}{*}{$p$ value (Dice)}} & \multicolumn{13}{c}{Dice of Each Class}                                                                                                                                                                                                              \\
                                   &                                       &                                        & \multicolumn{1}{l|}{}                                  & Sp               & RK               & LK               & Ga               & Es               & Li               & St               & Ao               & IVC              & PSV              & PA               & RAG              & LAG              \\ \hline
V-Net (fully)                      & $62.09_{\pm1.2}$                      & $10.28_{\pm3.9}$                       & $<0.01$                                                & 84.6             & 77.2             & 73.8             & 73.3             & 38.2             & 94.6             & 68.4             & 72.1             & 71.2             & 58.2             & 48.5             & 17.9             & 29.0             \\
Sup Only                           & $23.99_{\pm7.3}$                      & $74.26_{\pm5.9}$                       & $<0.01$                                                & 25.0             & 27.1             & 9.3              & 4.4              & 43.4             & 85.2             & 54.0             & 48.8             & 14.7             & 0.1              & 0.0              & 0.0              & 0.0              \\
UA-MT~\citep{yu2019uncertainty}    & $20.26_{\pm2.2}$                      & $71.67_{\pm7.4}$                       & $<0.01$                                                & 48.2             & 31.7             & 22.2             & 0.0              & 0.0              & 81.2             & 29.1             & 23.3             & 27.5             & 0.0              & 0.0              & 0.0              & 0.0              \\
URPC~\citep{luo2022semiurpc}       & $25.68_{\pm5.1}$                      & $72.74_{\pm15.5}$                      & $<0.01$                                                & 66.7             & 38.2             & 56.8             & 0.0              & 0.0              & 85.3             & 33.9             & 33.1             & 14.8             & 0.0              & 5.1              & 0.0              & 0.0              \\
CPS~\citep{chen2021cps}            & $33.55_{\pm3.7}$                      & $41.21_{\pm9.1}$                       & $<0.01$                                                & 62.8             & 55.2             & 45.4             & 35.9             & 0.0              & 91.1             & 31.3             & 41.9             & 49.2             & 8.8              & 14.5             & 0.0              & 0.0              \\
SS-Net~\cite{wu2022exploring}      & $35.08_{\pm2.8}$                      & $50.81_{\pm6.5}$                       & $<0.01$                                                & 62.7             & 67.9             & 60.9             & 34.3             & 0.0              & 89.9             & 20.9             & 61.7             & 44.8             & 0.0              & 8.7              & 4.2              & 0.0              \\
CLD~\cite{lin2022calibrating}      & $41.07_{\pm1.2}$                      & $32.15_{\pm3.3}$                       & $<0.01$                                                & 62.0             & 66.0             & 59.3             & 61.5             & 0.0              & 89.0             & 31.7             & 62.8             & 49.4             & 28.6             & 18.5             & 0.0              & 5.0              \\
DHC~\citep{wang2023dhc}            & $48.61_{\pm0.9}$                      & $10.71_{\pm2.6}$                       & $<0.01$                                                & 62.8             & 69.5             & 59.2             & \underline{66.0} & 13.2             & 85.2             & 36.9             & 67.9             & 61.5             & 37.0             & 30.9             & 31.4             & 10.6             \\
AllSpark~\citep{wang2024allspark}  & $60.68_{\pm0.6}$                      & $2.37_{\pm0.3}$                        & $<0.01$                                                & 86.3             & \underline{79.6} & 77.8             & 60.4             & 60.7             & 92.3             & 63.7             & 75.0             & 69.9             & \underline{60.2} & \underline{57.7} & 0.0              & 5.2              \\
GenericSSL~\citep{wang2023towards} & $60.88_{\pm0.7}$                      & $2.52_{\pm0.4}$                        & $<0.01$                                                & 85.2             & 66.9             & 67.0             & 52.7             & 62.9             & 89.6             & 52.1             & 83.0             & 74.9             & 41.8             & 43.4             & \underline{44.8} & 27.2             \\
S\&D-Messenger~\citep{zhang2025s}  & $\underline{68.38}_{\pm0.6}$          & $\underline{2.16}_{\pm0.6}$            & $<0.01$                                                & \underline{88.9} & \textbf{86.2}    & \textbf{86.6}    & 45.1             & \underline{66.3} & \underline{94.4} & \textbf{73.6}    & \underline{83.0} & \underline{75.3} & \textbf{60.5}    & 55.1             & 34.1             & \underline{39.8} \\
\textbf{BCMDA (ours)}              & $\textbf{71.20}_{\pm0.4}$             & $\textbf{1.65}_{\pm0.5}$               & -                                                      & \textbf{92.8}    & 77.4             & \underline{81.6} & \textbf{67.1}    & \textbf{68.6}    & \textbf{95.1}    & \underline{69.3} & \textbf{87.4}    & \textbf{79.4}    & 59.7             & \textbf{58.4}    & \textbf{43.1}    & \textbf{45.7}    \\ \hline
\end{tabular}
}
\end{table}

\begin{figure}[!t]
\centerline{\includegraphics[width=1.0\columnwidth]{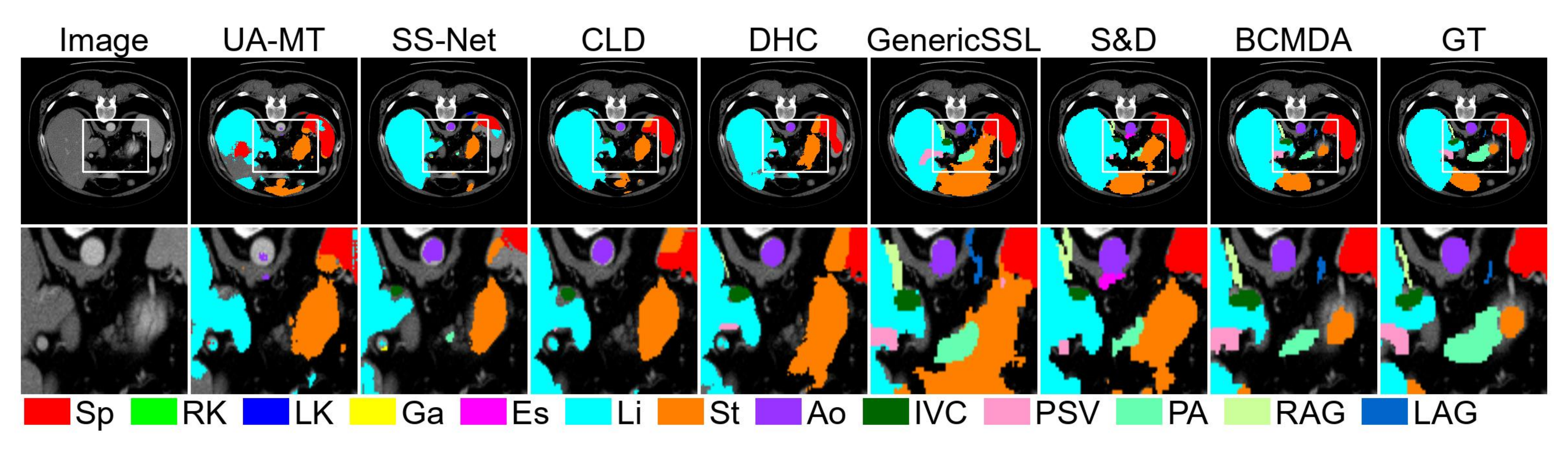}}
\vspace{-4mm}
\caption{Visualization results on the Synapse dataset.}
\label{Synapse}
\end{figure}

\subsubsection{Synapse dataset}
\label{IV-B4}
Table~\ref{tabsyn} presents the quantitative results compared with other methods using 20\% labeled data on the Synapse dataset. Notably, our method, BCMDA, was not specifically designed for this task, yet it achieves SOTA performance with an average Dice score of 71.20\% and a significantly lower ASD of 1.65\,mm, demonstrating superior boundary precision. Compared to recent competitive frameworks, BCMDA outperforms S\&D-Messenger at 68.38\%, GenericSSL at 60.88\%, and DHC at 48.61\% by substantial margins. In class-specific metrics, it attains the highest Dice score for several organs, notably the spleen at 92.8\% and gallbladder at 67.1\%. While S\&D-Messenger shows strong performance in large organs like the stomach at 73.6\%, it struggles with smaller or highly variable structures like the gallbladder at 45.1\%, where BCMDA provides a 22\% improvement. Similarly, while GenericSSL and DHC maintain reasonable average scores, they often fail to capture fine-grained details, resulting in higher ASD values of 2.52\,mm and 10.71\,mm respectively, compared to our much lower 1.65\,mm. Statistical analysis indicates that all p-values are less than 0.01, demonstrating that the improvements are statistically significant. Qualitative results in Fig.~\ref{Synapse} further illustrate that our method produces more complete and anatomically consistent masks, preserving structural integrity even in challenging regions. A visual comparison reveals that DHC often produces fragmented segments with significant noise, and GenericSSL tends to over-segment or merge adjacent organ boundaries. While S\&D-Messenger produces cleaner results, it still exhibits ``island'' artifacts and fails to accurately delineate the complex contours of the liver (Li) and  left adrenal gland (LAG). In contrast, BCMDA generates predictions that align most closely with the Ground Truth, effectively suppressing false positives and maintaining spatial continuity. This confirms that our method not only achieves higher numerical accuracy but also better understands the underlying spatial relationships between organs, even under limited supervision.

\subsection{Ablation analysis}
\label{IV-C}
We primarily conducted ablation studies on the Fundus dataset to evaluate the effectiveness of each component and hyperparameter in the proposed method. Except for the modified components, all other experimental settings remained the same.

\begin{figure}[!t]
\centerline{\includegraphics[width=\columnwidth]{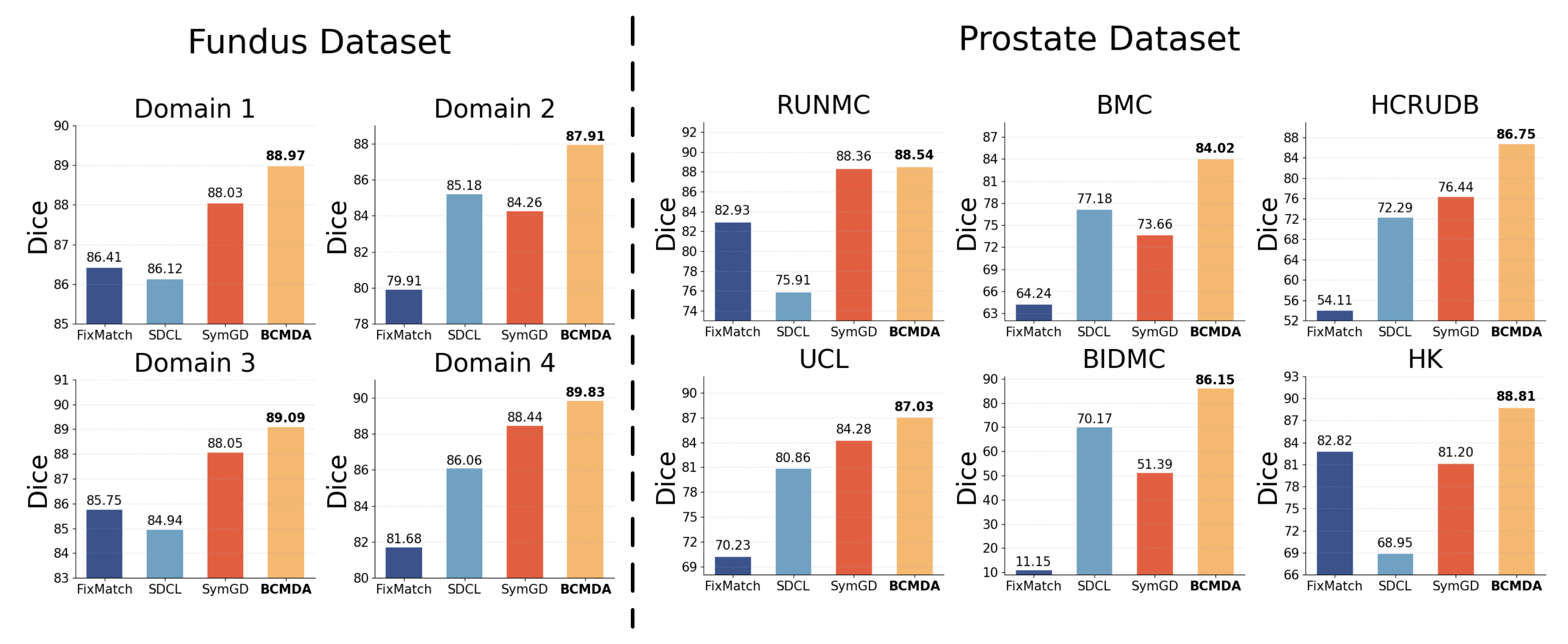}}
\vspace{-4mm}
\caption{Performance visualization comparison of BCMDA with other methods on the Fundus and Prostate datasets using 20 labeled data.}
\label{visualcom}
\end{figure}

\begin{table}[!t]
\centering
\caption{
Ablation study results on Fundus dataset.
}
\vspace{+4mm}
\label{tab4}
\resizebox{\columnwidth}{!}{
\begin{tabular}{c|cccccccc|cccc|cccc}
\toprule
\multirow{2}{*}{Method} &
  \multirow{2}{*}{BCMix} &
  \multirow{2}{*}{FixMix} &
  \multirow{2}{*}{PDMix} &
  \multirow{2}{*}{FDA} &
  \multirow{2}{*}{AVG} &
  \multirow{2}{*}{PA} &
  \multirow{2}{*}{BPA} &
  \multirow{2}{*}{PPLC} &
  \multicolumn{4}{c|}{(Optic Cup / Optic Disc Segmentation) Dice $\uparrow$} &
  Dice $\uparrow$ &
  Jaccard $\uparrow$ &
  95HD $\downarrow$ &
  ASD $\downarrow$ \\
 &
   &
   &
   &
   &
   &
   &
   &
   &
  Domain 1 &
  Domain 2 &
  Domain 3 &
  Domain 4 &
  \multicolumn{4}{c}{Avg.} \\ \hline
\#1 &
  \checkmark &
   &
   &
   &
   &
   &
   &
   &
  82.76 / 92.51 &
  73.75 / 89.50 &
  81.91 / 92.62 &
  81.73 / 93.09 &
  85.98 &
  76.85 &
  9.41 &
  4.44 \\
\#2 &
  \checkmark &
  \checkmark &
   &
   &
   &
   &
   &
   &
  82.91 / 93.48 &
  79.66 / 91.66 &
  83.49 / 93.19 &
  84.38 / 93.54 &
  87.79 &
  79.33 &
  7.76 &
  3.70 \\
\#3 &
  \checkmark &
   &
  \checkmark &
   &
   &
   &
   &
   &
  82.68 / 93.35 &
  79.41 / 90.92 &
  83.30 / 93.18 &
  83.46 / 93.83 &
  87.51 &
  78.95 &
  7.86 &
  3.78 \\
\#4 &
  \checkmark &
  \checkmark &
  \checkmark &
   &
   &
   &
   &
   &
  83.76 / 93.52 &
  81.52 / 91.83 &
  83.96 / 93.60 &
  84.60 / \underline{94.14} &
  88.37 &
  80.12 &
  7.56 &
  3.59 \\
\#5 &
  \checkmark &
  \checkmark &
  \checkmark &
  \checkmark &
  \multicolumn{1}{l}{} &
  \multicolumn{1}{l}{} &
  \multicolumn{1}{l}{} &
  \multicolumn{1}{l|}{} &
  81.82 / 93.39 &
  80.74 / 90.22 &
  83.54 / 93.29 &
  83.32 / 93.52 &
  87.48 &
  78.77 &
  7.83 &
  3.81 \\
\#6 &
  \checkmark &
  \checkmark &
  \checkmark &
  \multicolumn{1}{l}{} &
  \checkmark &
  \multicolumn{1}{l}{} &
  \multicolumn{1}{l}{} &
  \multicolumn{1}{l|}{} &
  83.51 / \textbf{93.94} &
  82.21 / 91.89 &
  \underline{84.27} / 93.04 &
  84.65 / 94.03 &
  88.44 &
  80.25 &
  \underline{7.32} &
  \underline{3.52} \\
\#7 &
  \checkmark &
  \checkmark &
  \checkmark &
   &
  \checkmark &
  \checkmark &
   &
   &
  83.18 / \underline{93.88} &
  82.41 / 91.84 &
  83.44 / \textbf{93.65} &
  84.91 / \textbf{94.36} &
  88.46 &
  80.29 &
  7.39 &
  3.56 \\
\#8 &
  \checkmark &
  \checkmark &
  \checkmark &
   &
  \checkmark &
   &
  \checkmark &
   &
  \underline{83.90} / 93.85 &
  \underline{82.83} / \underline{92.15} &
  83.93 / \underline{93.58} &
  \underline{84.92} / 93.85 &
  \underline{88.63} &
  \underline{80.57} &
  7.38 &
  3.54 \\ \hline
\#9 &
  \checkmark &
  \checkmark &
  \checkmark &
   &
  \checkmark &
   &
  \checkmark &
  \checkmark &
  \textbf{84.25} / 93.68 &
  \textbf{83.35} / \textbf{92.46} &
  \textbf{84.67} / 93.51 &
  \textbf{85.53} / 94.12 &
  \textbf{88.95} &
  \textbf{81.03} &
  \textbf{7.21} &
  \textbf{3.45} \\ \bottomrule
\end{tabular}
}
\end{table}

\subsubsection{Effectiveness of each component}
\label{IV-C1}
To evaluate the contribution of each component, we conducted ablation experiments on the Fundus dataset using 20 labeled samples. As shown in Table~\ref{tab4}, \textbf{BCMix} denotes the use of the method between labeled and unlabeled data, serving as the baseline upon which the subsequent components are built. 
\textbf{FDA} denotes a variant in which our image generation strategy is replaced by the Fourier domain adaptation~\citep{yang2020fda} method. \textbf{AVG} represents the use of averaged predictions from both virtual and real unlabeled data as pseudo labels. \textbf{PA} indicates that only one prototype per class is maintained throughout training. In the absence of \textbf{PPLC}, pseudo labels are generated by the Linear classifier. Employing either \textbf{FixMix} or \textbf{PDMix} individually already results in marked performance improvements, while the combination of both strategies leads to further gains. In contrast, replacing our method with \textbf{FDA} yields only marginal improvements over the baseline, highlighting the superiority of our approach. The addition of \textbf{AVG} brings modest performance gains, while the introduction of \textbf{PA} shows negligible impact. Conversely, incorporating \textbf{BPA} leads to further improvements. With the integration of \textbf{PPLC}, our framework ultimately achieves the best overall performance.

\subsubsection{Impacts of hyper-parameters}
\label{IV-C2}
\begin{figure}[!t]
\centerline{\includegraphics[width=\columnwidth]{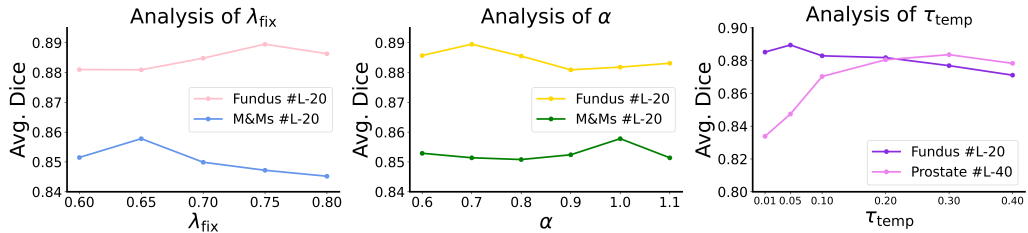}}
\vspace{-4mm}
\caption{Hyper-parameter analysis of $\lambda_{\text{fix}}$, $\alpha$, and $\tau_{\text{temp}}$.}
\label{hyper_lamda}
\end{figure}

In our framework, \( W' \), \( \lambda_{\text{fix}} \), \( \alpha \), and \( \tau_{\text{temp}} \) are four critical hyperparameters. We conducted ablation studies to analyze their effects. As shown in Table~\ref{tab5}, varying the scale of \( W' \) has minimal impact on the performance. Fig.~\ref{hyper_lamda} presents the analysis of \( \lambda_{\text{fix}} \) and \( \alpha \). The results indicate that \( \alpha \) has a relatively minor effect on performance. Moreover, both excessively large and small values of \( \lambda_{\text{fix}} \) lead to suboptimal outcomes, suggesting that appropriate control of the mixing ratio in the generated images is essential for constructing well-aligned virtual domain data. From Fig.~\ref{hyper_lamda}, we observe that \( \tau_{\text{temp}} \) significantly influences model performance. Specifically, for the more challenging Prostate dataset, a lower value of \( \tau_{\text{temp}} \) may result in overconfident predictions and degraded performance. In contrast, for the more easily converged Fundus dataset, a smaller \( \tau_{\text{temp}} \) value can be adopted.

\begin{table}[!t]
\centering
\caption{
Comparison with different $W'$.
}
\vspace{+4mm}
\label{tab5}
\begin{tabular}{l|cccc}
\toprule
\multicolumn{1}{c|}{\multirow{2}{*}{$W'$}} &
  \multicolumn{1}{l}{Dice $\uparrow$} &
  \multicolumn{1}{l}{Jaccard $\uparrow$} &
  \multicolumn{1}{l}{95HD $\downarrow$} &
  \multicolumn{1}{l}{ASD $\downarrow$} \\
\multicolumn{1}{c|}{} & \multicolumn{4}{c}{Avg.}    \\  \hline
$\frac{1}{2}W$               & 88.84 & 80.86 & \underline{7.27} & 3.47 \\
$\frac{1}{4}W$               & \textbf{88.95} & \textbf{81.03} & \textbf{7.21} & \textbf{3.45} \\
$\frac{1}{8}W$               & \underline{88.90} & \underline{80.93} & 7.38 & \underline{3.45} \\ \bottomrule
\end{tabular}
\end{table}

\subsubsection{Influence of PPLC}
\label{IV-C3}
\begin{figure}[!t]
\begin{minipage}{0.40\columnwidth}
\centerline{\includegraphics[width=\columnwidth]{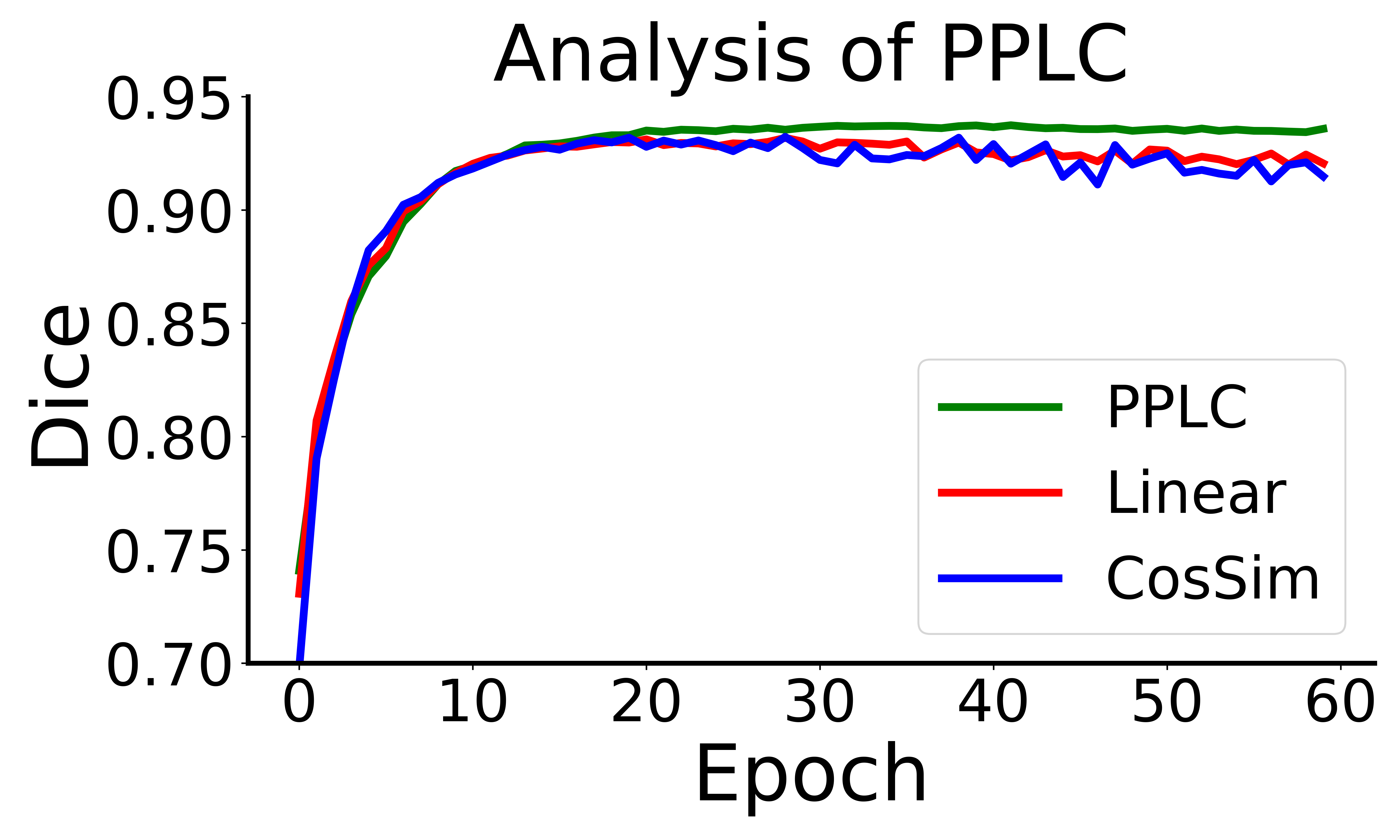}}
\vspace{-4mm}
\caption{Quantitative analysis of pseudo labels quality.}
\label{PPLC}
\end{minipage}
\hfill
\begin{minipage}{0.58\columnwidth}
\centerline{\includegraphics[width=\columnwidth]{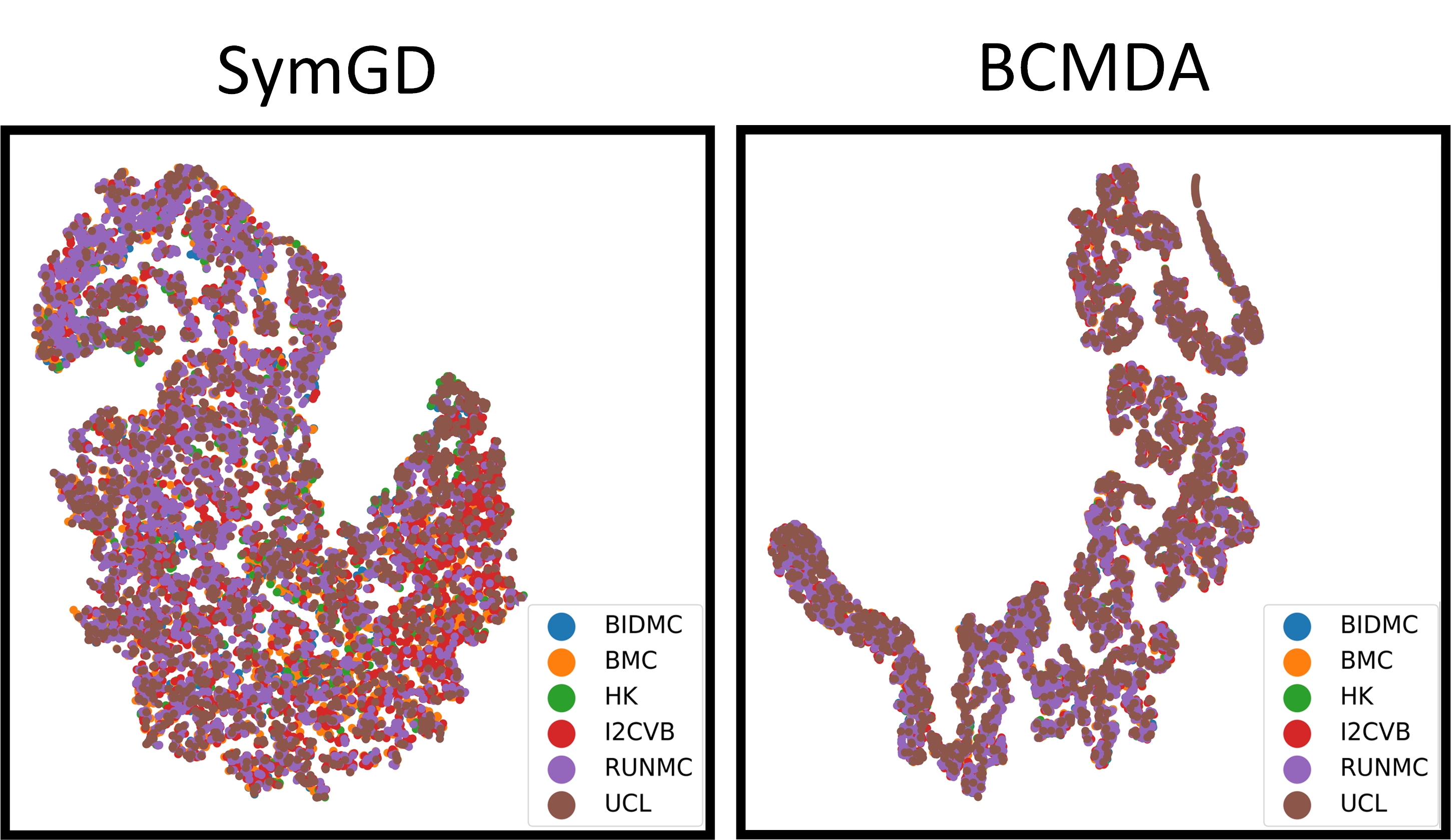}}
\vspace{-4mm}
\caption{Visualization of feature maps for foreground class across various domains using t-SNE.}
\label{tsne}
\end{minipage}
\end{figure}

We evaluated the effectiveness of PPLC on the Fundus dataset using 20 labeled samples, with the quality of pseudo labels quantitatively assessed by their Dice scores, as illustrated in Fig.~\ref{PPLC}. 
The green, red, and blue curves represent the predictions refined by PPLC, predictions using only the Linear classifier, and predictions using only the CosSim classifier as supervision signals, respectively. It is evident that the pseudo labels generated by PPLC demonstrate significantly higher quality than those from the other two classifiers, while also exhibiting superior stability.

\subsubsection{t-SNE visualization}
\label{IV-C4}
To further demonstrate the advantages of our method, we performed t-SNE visualization on the feature maps extracted by the backbone, i.e., the last-layer features before the classifiers, trained with 40 labeled samples from the Prostate dataset, as shown in Fig.~\ref{tsne}. Compared with SymGD, the feature manifold learned by our method is more compact, exhibiting tighter intra-class clustering, which indicates fewer isolated features and more discriminative representations across different classes. As a result, clearer decision boundaries are formed for the classifiers, leading to improved classification performance.

\begin{table}[!t]
\centering
\caption{
Comparison of performance and complexity for different methods
}
\vspace{+4mm}
\label{tab_complex}
\begin{tabular}{c|cc|ccc}
\hline
\multirow{2}{*}{Method} & \multicolumn{2}{c|}{Performance} & \multicolumn{3}{c}{Complexity}       \\ \cline{2-6} 
                        & Dice$\uparrow$ & ASD$\downarrow$ & Epoch(s) & Iteration(s) & Memory(GB) \\ \hline
SymGD                   & 87.20          & 3.92            & 101.31   & 0.20         & \textbf{4.07}       \\
\textbf{BCMDA}                   & \textbf{88.95}          & \textbf{3.45}            & \textbf{70.88}    & \textbf{0.14}         & 4.79       \\ \hline
\end{tabular}
\end{table}

\subsubsection{Complexity Analysis}
\label{IV-C6}
Table~\ref{tab_complex} presents a comparison of computational complexity between our method and SymGD, including both training time and memory usage, as well as performance. It can be observed that although our method consumes relatively more memory, the training time is actually shorter, and the performance is significantly improved. We attribute the peak memory usage to the Softmax operation in the computation of the bidirectional correlation maps. Conversely, while SymGD uses less memory, its reliance on Fast Fourier transform operations increases the number of computations, resulting in longer training time. Overall, the additional memory consumption of our method is acceptable, given the substantial gains in computational efficiency and performance.

\subsubsection{Limitations and future works}
\label{IV-C7}
\begin{figure}[!t]
\centerline{\includegraphics[width=\columnwidth]{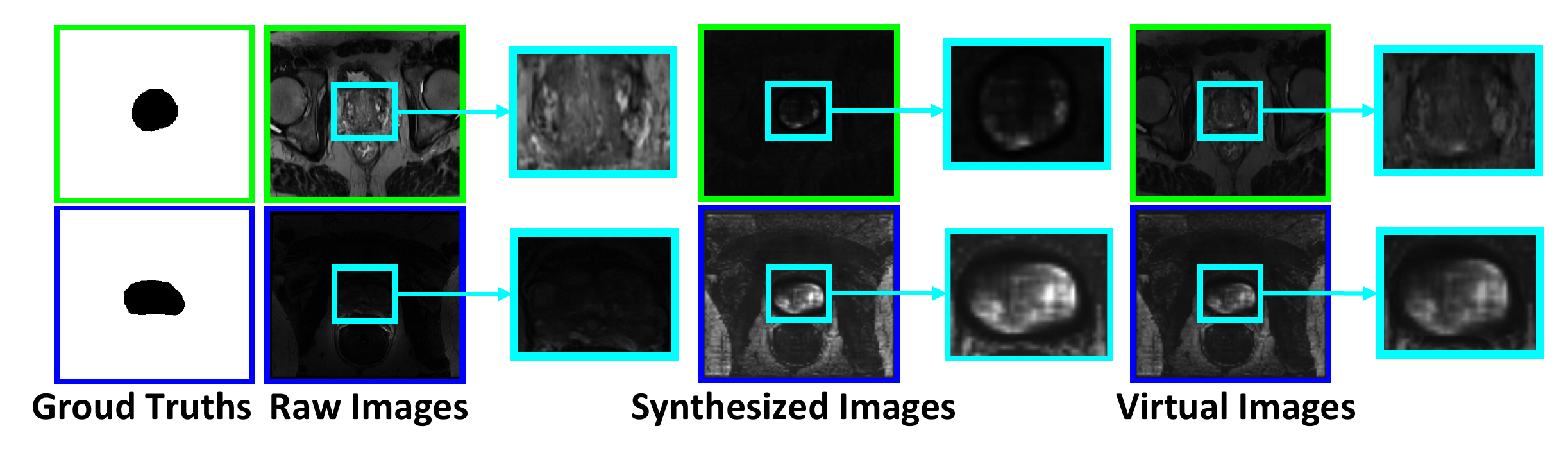}}
\vspace{-4mm}
\caption{This figure illustrates generated virtual images for challenging samples where misinterpretation by the model occurs.}
\label{limitvis}
\end{figure}

Although we only directly use the feature maps output by the backbone to guide the image synthesis, after applying MixUp with the original image, not only is the data distribution aligned, but it also compensates for the texture details, demonstrating excellent performance. However, our method still presents certain limitations, as illustrated in Fig.~\ref{limitvis}, which shows virtual images generated for challenging samples where the model tends to misinterpret pixel semantics. In the figure, the model learns correctly from the labeled image with a green border, while it tends to misinterpret pixel semantics in the unlabeled image with a blue border. Since our image synthesis process relies on inter-image correlations, accurately understanding both the raw image pixels and the source pixels used for synthesis is crucial. The synthesized labeled image generally maintains a correct foreground structure but lacks texture details, possibly due to misinterpretation of the source pixels. Nevertheless, the corresponding raw image contains rich textures, which can effectively compensate for the texture details of the synthesized image using MixUp. By contrast, due to misinterpretation of its own pixels and limited texture details, the virtual unlabeled image exhibits a foreground structure inconsistent with reality and displays muddy textures. It is apparent that the blurred textures and semantic errors in the structure of these images will affect the model's recognition and segmentation results. Therefore, in future work, we plan to investigate additional modules aimed at generating higher-quality feature maps and refining the image synthesis process. These enhancements are expected to improve the robustness and performance of BCMDA, thereby making it a more reliable solution for real-world clinical applications. 

\section{Conclusion}
In this paper, we first analyze the issues of domain shift and confirmation bias in the MiDSS scenario, as well as the shortcomings of current SSMS and UDA methods. Based on this analysis, we propose a novel MiDSS framework, BCMDA. Specifically, its innovation lies in the use of bidirectional correlation maps to generate virtual data that bridges the domain gap. Additionally, we introduce the learnable prototype cosine similarity classifiers to achieve bidirectional prototype alignment and correct confirmation bias. In contrast to other SSMS and UDA methods, BCMDA simultaneously provides solutions to the issues inherent in each of them. Experimental results demonstrate the effectiveness and superiority of our method, which maintains strong performance even under extremely limited labeled data. To further validate its generalizability, we conduct additional experiments on 3D dataset, confirming that BCMDA can be effectively extended to volumetric segmentation task. Furthermore, we plan to explore extending this method to multi-modal medical imaging in the future, investigating how bidirectional correlation maps can effectively facilitate domain adaptation and bridge the domain gap between different imaging modalities, such as CT and MRI.

\section*{CRediT authorship contribution statement} 
\textbf{Bentao Song}: Conceptualization, Validation, Formal analysis, Investigation, Resources, Data Curation, Writing - Original Draft, Visualization. \textbf{Jun Huang:} Writing - Review \& Editing, Investigation, Software. \textbf{Qingfeng Wang}: Writing - Review \& Editing, Methodology, Validation, Supervision, Funding acquisition.

\section*{Acknowledgement}
This work was supported by the construction project of the Fujiang
Laboratory Nuclear Medicine Artificial Intelligence Research Center (No.
2023ZYDF074).

\bibliographystyle{apalike} 
\bibliography{cas-refs.bib}

\end{document}